\documentclass[10pt,a4paper,logo]{paper}

\usepackage[all]{hypcap}

\usepackage[authoryear, round]{natbib}

\usepackage{nicefrac}       % compact symbols for 1/2, etc.
\usepackage{multirow}
\usepackage{cleveref}
\usepackage{dsfont}
\usepackage{makecell}
\usepackage{subcaption}
\usepackage{bbm}
\usepackage{wrapfig}
\usepackage{mathtools,xparse}
\usepackage{array}
\usepackage{arydshln}
\usepackage{scalerel}
\usepackage{tablefootnote}
\usepackage{tocloft}
\usepackage{adjustbox}
\usepackage{setspace}
\usepackage[most,skins,theorems]{tcolorbox}

\newtcolorbox{promptbox}{
  colback=gray!5,
  colframe=gray!70,
  boxrule=0.8pt,
  arc=2mm,
  left=10pt,
  right=10pt,
  top=10pt,
  bottom=10pt,
  width=\linewidth,
  fontupper=\ttfamily\footnotesize,
  enhanced,
}

\newboolean{showsection}
\setboolean{showsection}{true}
\makeatletter
\@namedef{ver@everyshi.sty}{}
\makeatother

\definecolor{custom_green}{rgb}{0.0, 0.5, 0.0}
\definecolor{custom_red}{rgb}{1.0, 0.01, 0.24}
\definecolor{custom_blue}{HTML}{C9DAF7}
\definecolor{custom_purple}{HTML}{D9D1E9}
\definecolor{title_blue}{HTML}{204899}
\definecolor{cite_blue}{HTML}{044dc1}
\definecolor{cite_purple}{HTML}{7406a7}
\newcommand{\cmark}{\color{custom_green}\ding{51}}
\newcommand{\xmark}{\color{custom_red}\ding{55}}

\hypersetup{
    colorlinks = true,
    citecolor = {cite_blue},
    linkcolor = {cite_purple},
    urlcolor = {cite_purple},
}

\definecolor{blanchedalmond}{rgb}{1.0, 0.92, 0.8}
\definecolor{carmine}{rgb}{0.59, 0.0, 0.09}
\definecolor{lightblue}{rgb}{0.22,0.45,0.70}%

\renewcommand{\mathbf}{\boldsymbol}

\makeatletter
\def\Ddots{\mathinner{\mkern1mu\raise\p@
\vbox{\kern7\p@\hbox{.}}\mkern2mu
\raise4\p@\hbox{.}\mkern2mu\raise7\p@\hbox{.}\mkern1mu}}
\makeatother

\numberwithin{equation}{section}

\definecolor{amaranth}{rgb}{0.9, 0.17, 0.31}
\definecolor{antiquebrass}{rgb}{0.8, 0.58, 0.46}
\definecolor{antiquefuchsia}{rgb}{0.57, 0.36, 0.51}
\definecolor{chromeyellow}{rgb}{0.31, 0.47, 0.26}

\newcommand{\1}{\mathds 1}

\usepackage{amsmath,amsfonts,bm}

\def\eqref#1{equation~\ref{#1}}
\def\1{\bm{1}}

\DeclareMathAlphabet{\mathsfit}{\encodingdefault}{\sfdefault}{m}{sl}
\SetMathAlphabet{\mathsfit}{bold}{\encodingdefault}{\sfdefault}{bx}{n}

\makeatletter
\def\mathcolor#1#{\@mathcolor{#1}}
\def\@mathcolor#1#2#3{%
  \protect\leavevmode
  \begingroup
    \color#1{#2}#3%
  \endgroup
}
\makeatother

\Crefformat{equation}{#2Eq.\;(#1)#3}
\Crefformat{figure}{#2Figure #1#3}
\Crefformat{assumption}{#2Assumption #1#3}
\Crefname{assumption}{Assumption}{Assumptions}

\usepackage{crossreftools}
\makeatletter
\renewcommand\footnoterule{%
  \kern 15\p@
  \hrule \@width 2in \kern 2.6\p@
  \vspace{4pt}
}
\makeatother
\newcommand{\method}{H2RB}

\title{Human-to-Robot Transfer via Bridging Representation and Action Space in VLAs}

\reportnumber{}
\correspondingauthor{Correspondence to: Jinwoo Shin (\email{jinwoos@kaist.ac.kr})}

\author[1]{Junhyeong Park}
\author[1]{Wook Jung}
\author[1, 2]{Taeyoung Kim}
\author[1, 2]{Huiwon Jang}
\author[1, 2]{Jinwoo Shin}

\affil[1]{KAIST}
\affil[2]{RLWRLD}

\begin{abstract}
    Pretrained Vision-Language-Action (VLA) policies provide a strong starting point for deployment, but adapting them to a target robot and environment still requires costly embodiment-specific demonstrations. 
    Human demonstrations are far easier to collect, yet directly leveraging them is challenging due to the substantial human-robot embodiment gap, particularly when the underlying VLA has been pretrained primarily on robot data. 
    We introduce \textbf{\method{}} (\textbf{H}uman-\textbf{T}o-\textbf{R}obot \textbf{B}ridge), a human-robot co-training framework for adapting pretrained VLA policies from unpaired human and robot demonstrations. 
    \method{} bridges the embodiment gap at both interfaces:
    geometry-anchored alignment grounds human and robot observations in a compatible
    representation, while a decoupled-decoder tokenizer shares action latents without
    forcing both domains through a single action mapping.
    This design allows human demonstrations to provide useful behavioral supervision while retaining the rich visual, semantic, and control priors of the pretrained VLA.
    Across real-world tasks, \method{} substantially improves object transfer performance from 8.0 to 51.8 and enables skill transfer to score 62.7, in settings where co-train baselines totally fail to transfer human-demonstrated skills.
    The benefit also persists under object-placement shifts, improving out-of-distribution transfer from 6.0 to 45.3.
    \method{} suggests a way for human demonstrations to serve as a scalable supervision source for adapting pretrained robotic policies, reducing the dependence on costly target-robot data collection.
\end{abstract}

\begin{document}

\maketitle

\section{Introduction}
\label{sec:intro}

Pretrained robotic policies, \textit{e.g.}, Vision-Language-Action (VLA) models 
\citep{brohan2023rt,kim2024openvla}, provide a strong starting point for adapting general robotic capabilities to a target deployment environment. 
However, a major bottleneck remains in the collection of target robot demonstrations, as collecting such data on physical robots is costly, time-consuming, and difficult to scale.
It requires access to the physical platform, a data collection pipeline (e.g., teleoperation), and repeated environment resets \citep{brohan2022rt,khazatsky2024droid}. 

Human demonstrations, in contrast, can be collected quickly and naturally without relying on dedicated teleoperation systems~\citep{chi2024universal,qiu2025humanoid,yang2025egovla,
wang2023mimicplay}, using only a lightweight camera setup. 
Yet, directly leveraging them to adapt robot policies remains challenging due to the substantial embodiment gap between humans and robots, particularly when the pretrained policy has been trained primarily on robot data.

Prior works bridge human and robot demonstrations by placing them in a common action space, either through latent actions inferred from frame transitions~\citep{ye2025latent,chen2026villa,dai2026conla,bu2025univla,chen2025moto} or through reconstructed world-space hand motion~\citep{li2025scalable,fu2026egograsp,hoque2026egodex}.
While these approaches reduce the action-space mismatch across embodiments, they leave the perceptual gap largely unresolved. 
As a result, VLAs must map distinct human and robot observations into a shared action space, making reliable robot action prediction difficult.
However, a common action space alone does not make human demonstrations compatible with a robot-pretrained VLA. Embodiment-dependent observations remain perceptually misaligned, while human and robot motions can still require different action realization dynamics.

\begin{figure*}[t]
    \centering
    \includegraphics[width=\linewidth]{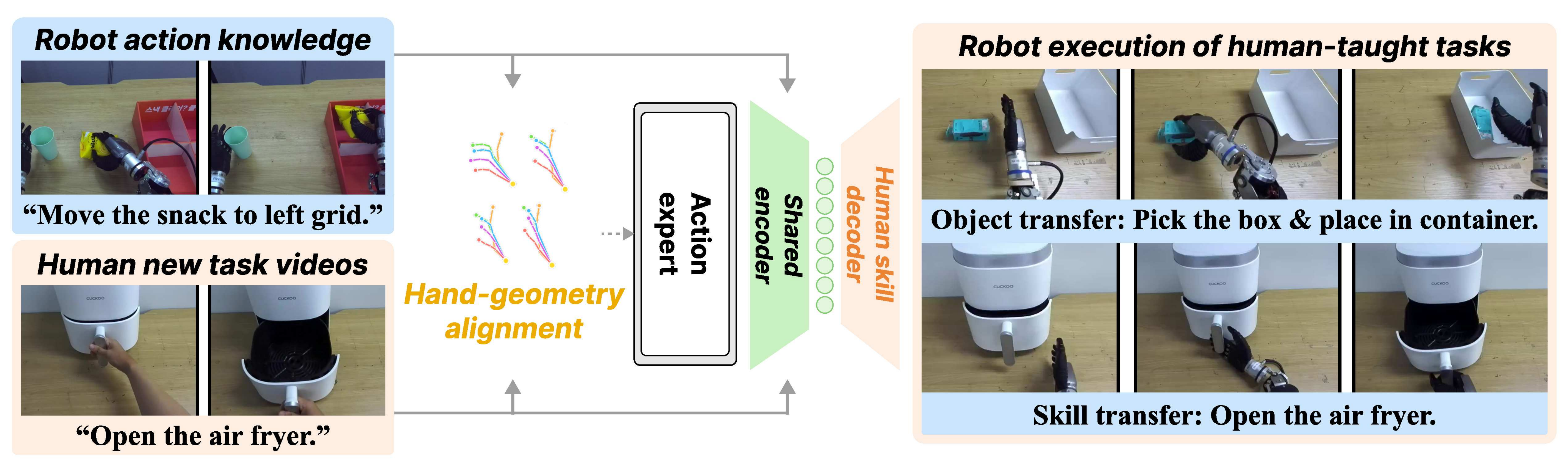}
    \vspace{-6mm}
    \caption{
        \textbf{Human-to-robot task transfer with \method{}.}
\method{} transfers task experience from human videos to robots using pretrained VLAs by bridging the cross-embodiment perception and action gaps, without target-robot demonstrations or paired demonstrations.
    }
    \label{fig:teaser}
    \vspace{-6mm}
\end{figure*}

\textbf{Our approach.}~
Based on this insight, we introduce \method{}, a training framework that incorporates human demonstrations into pretrained VLA policies without paired human-robot trajectories.
First, a geometry-anchored alignment uses wrist-fingertip geometry to ground VLM output features across embodiments, without paired demonstrations or human-to-robot observation synthesis. 
Second, a decoupled-decoder latent tokenizer shares action latents across embodiments while using separate decoders to avoid collapsing human and robot motion dynamics into a single action mapping.
Together, these components enable human-to-robot task transfer while facilitating learning from heterogeneous human-robot experience, as illustrated in~\Cref{fig:teaser}.

We evaluate \method{} on two complementary forms of human-to-robot adaptation: within-skill object transfer and skill-level transfer. On OpenArm, where target behaviors are provided only through human demonstrations, \method{} improves object transfer from 8.0 to 51.8 and achieves a skill-transfer score of 62.7, where shared-action co-training baselines fail. Heterogeneous human-robot training across multiple embodiments further improves these scores to 57.2 and 81.7, respectively.

We additionally study robot-only downstream adaptation on RB-Y1, where cross-embodiment alignment improves success from 72\% to 78\% under matched heterogeneous training. Together, these results suggest that learning action-relevant representations from heterogeneous experience benefits both human-to-robot transfer and downstream robot adaptation.
\section{Method}
\label{sec:method}

\begin{figure}[!t]
    \centering
    \captionsetup[subfigure]{}
    % ---------------- (a) ----------------
    ~
    \begin{subfigure}[t]{0.35\linewidth}
        \centering
        \includegraphics[width=\linewidth]{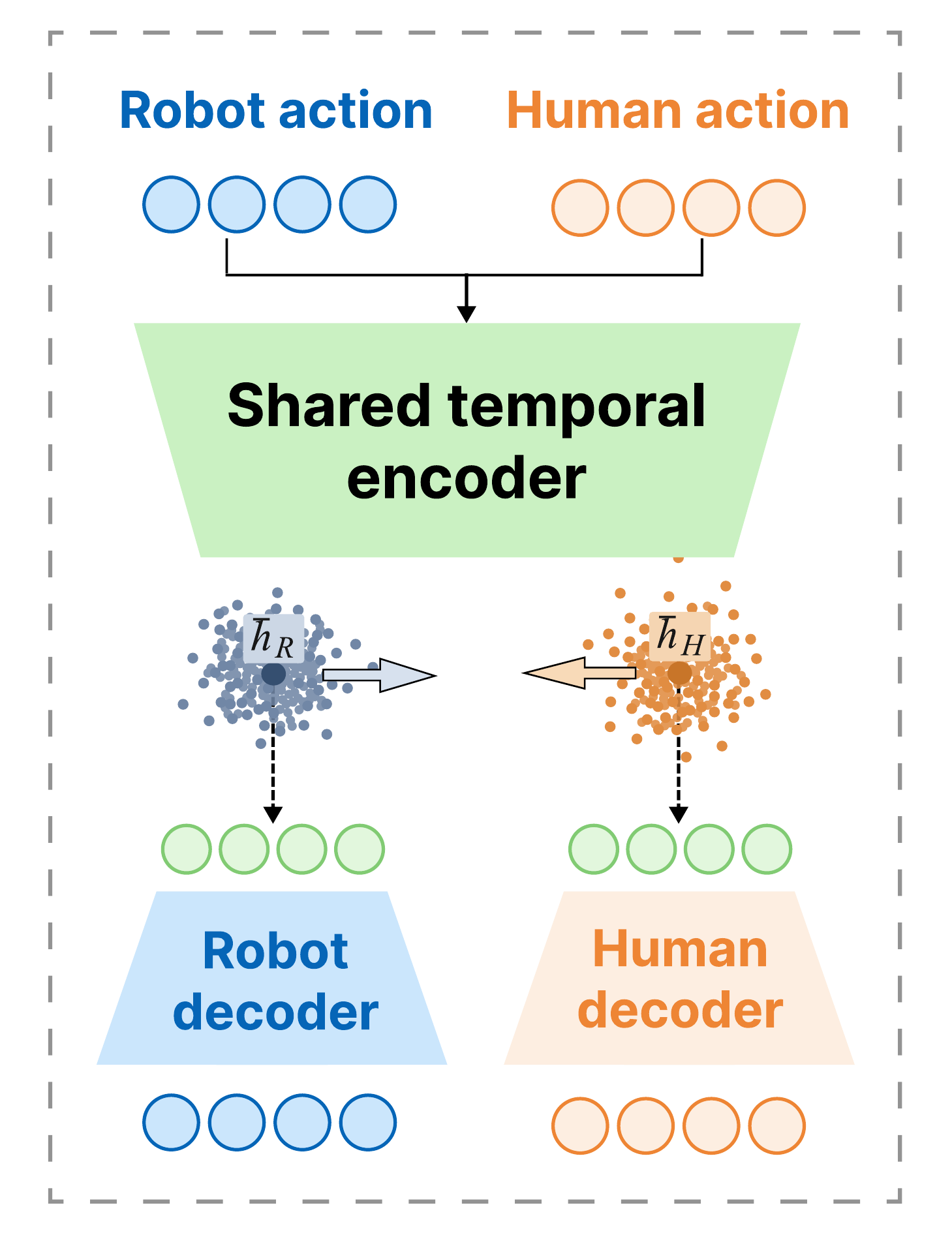}
        \captionsetup{width=1.05\linewidth}
        \caption{Decoupled-decoder latent tokenizer}
        \label{fig:pipeline-a}
    \end{subfigure}
    \hfill
    % ---------------- (b) ----------------
    \begin{subfigure}[t]{0.63\linewidth}
        \centering
        \includegraphics[width=\linewidth]{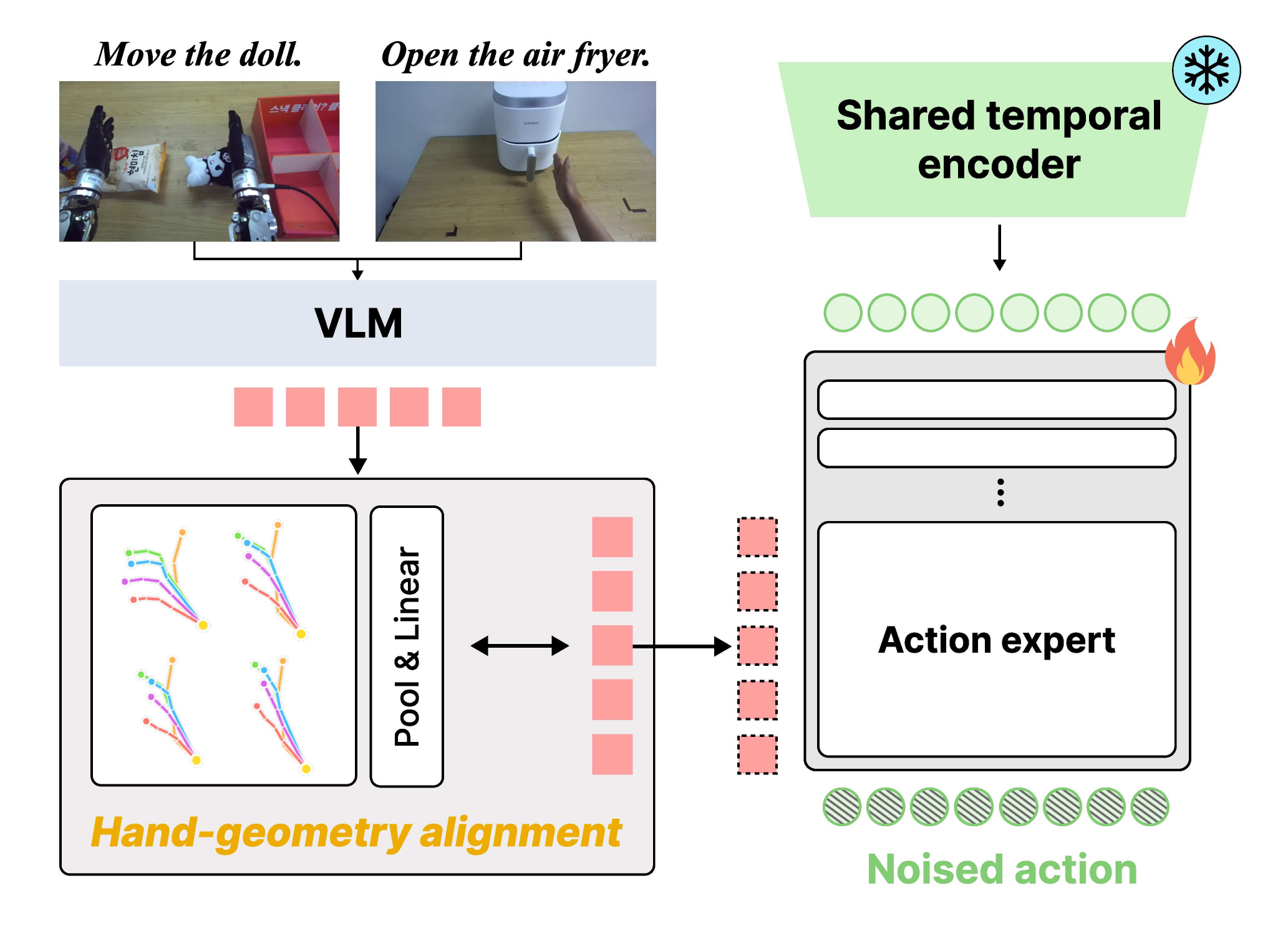}

        \caption{Hand geometry-anchored alignment}
        \label{fig:pipeline-b}
    \end{subfigure}

    \caption{
        \textbf{Overview of \method{}.}
        \textbf{(a)} A shared temporal encoder maps human and robot action chunks
        into a common latent action space, while embodiment-specific decoders
        reconstruct actions for each domain.
        \textbf{(b)} Hand geometry-anchored alignment grounds VLA conditioning across embodiments, while the frozen encoder provides latent action targets.
    }
    \label{fig:pipeline}
\end{figure}

\method{} bridges human-robot gaps through shared latent actions with embodiment-specific decoders and geometry-grounded VLM features, as illustrated in \Cref{fig:pipeline}.

\subsection{Preliminaries: Vision-Language-Action (VLA) models}
\label{sec:preliminaries}
At timestep $t$, a VLA predicts a $K$-step action chunk
$\mathbf A_t :=
[\mathbf a_t,\mathbf a_{t+1},\ldots,\mathbf a_{t+K-1}]$
from a visual observation $\mathbf o_t$, a language instruction $\mathbf c$,
and the proprioceptive state $\mathbf s_t$.
Let $\mathbf h_t$ denote the VLM conditioning tokens obtained from
$(\mathbf o_t,\mathbf c)$.
Following flow-based VLAs~\citep{bjorck2025gr00t,black2024pi_0,kim2026rldx}, we use conditional flow matching~\citep{lipman2022flow}: for Gaussian noise $\boldsymbol\epsilon\sim\mathcal N(\mathbf 0,\mathbf I)$ and flow time $\tau\in[0,1]$, we define 
% $\mathbf a_{t:t+K}^\tau=(1-\tau)\boldsymbol\epsilon+\tau\mathbf a_{t:t+K}$ 
$\mathbf A_{t,\tau}
=(1-\tau)\boldsymbol\epsilon+\tau\mathbf A_t$
and optimize 
$\mathcal L_{\mathrm{FM}}
=
\mathbb E[
\|\mathbf v_\theta(
\mathbf A_{t,\tau},
\tau
\mid
\mathbf s_t,\mathbf h_t)
-(\mathbf A_t-\boldsymbol\epsilon)
\|^2
].$
Here $\mathbf v_\theta$ is the velocity field predicted by the action expert.

\subsection{Learning a shared action space}
\label{sec:shared-action}
\label{sec:shared-action-dynamics}

To reduce the action gap between human and robot demonstrations, we first express their motions in a common physical parameterization. We then learn a shared latent prediction space while retaining separate decoders for embodiment-dependent trajectory structure.

\vspace{0.01in}
\noindent{\bf Shared wrist-hand parameterization.}
For human videos, we first use a HaWoR-based~\citep{zhang2025hawor} reconstruction pipeline to recover wrist motion and hand articulation.
We then transform the reconstructed motion into the robot coordinate convention and retarget it to the robot hand.
We express human and robot actions in a common wrist-hand representation~\citep{kim2026ego},
\textit{i.e.}, $\mathbf a_t =
[\mathbf p_t,\mathbf r_t,\mathbf q_t]$,
where $\mathbf p_t\in\mathbb R^3$ is the wrist position,
$\mathbf r_t\in\mathbb R^6$ is the 6D rotation representation,
and $\mathbf q_t \in \mathbb R^D$ is the hand configuration with $D$ joint angles for both human and robot demonstrations.
At deployment, we convert the predicted wrist pose into robot arm motion via inverse kinematics.

\vspace{0.01in}
\noindent{\bf Decoupled-decoder latent tokenizer.}
A shared wrist-hand parameterization does not eliminate source-dependent action differences induced by distinct human and robot perceptions.
We therefore learn a shared latent action space while separating decoders for each embodiment to capture these differences.
Specifically, given human actions $\mathbf{A}^H_t$ and robot actions $\mathbf{A}^R_t$ in a common wrist-hand representation, we train a variational action autoencoder~\citep{kingma2013auto} with a shared encoder and embodiment-specific decoders (see \Cref{fig:pipeline-a}).
For each embodiment $d\in\{H,R\}$, the encoder parameterizes the posterior $q_\psi(\mathbf Z_t^d\mid\mathbf A_t^d)$, from which we sample a latent action $\widetilde{\mathbf Z}_t^d$ and reconstruct the corresponding action using decoder $D_d$:

\begin{equation}
\widetilde{\mathbf Z}_t^d
\sim q_\psi(\cdot\mid\mathbf A_t^d),
\qquad
\widehat{\mathbf A}_t^d
=
D_d(\widetilde{\mathbf Z}_t^d)
\label{eq:latent-action}
\end{equation}

We optimize the tokenizer with reconstruction and KL objectives, and further
couple the two latent distributions through position-wise centroid matching
with variance and covariance regularization
(see Appendix~\ref{app:tokenizer_objectives}).
This encourages both sources to occupy a compatible latent space while
allowing each decoder to preserve source-specific trajectory structure.
After training the tokenizer, we freeze the encoder and both decoders, and use
$\mathbf Z_t^d
=
E_\psi(\mathbf A_t^d)$
as the VLA prediction target, where $E_\psi$ is the posterior-mean mapping of $q_\psi$.

We apply the flow matching objective introduced in \Cref{sec:preliminaries} to these latent targets and denote the resulting prediction loss by $\mathcal L_{\mathrm{pred}}$.

\subsection{Geometry-anchored VLM conditioning}
\label{sec:geometry-ground}
\label{sec:policy-deployment}

% ------------------------------------------------------------------
% Qualitative success rollouts
% ------------------------------------------------------------------
\begin{figure*}[t]
    \centering
    \includegraphics[width=\textwidth]{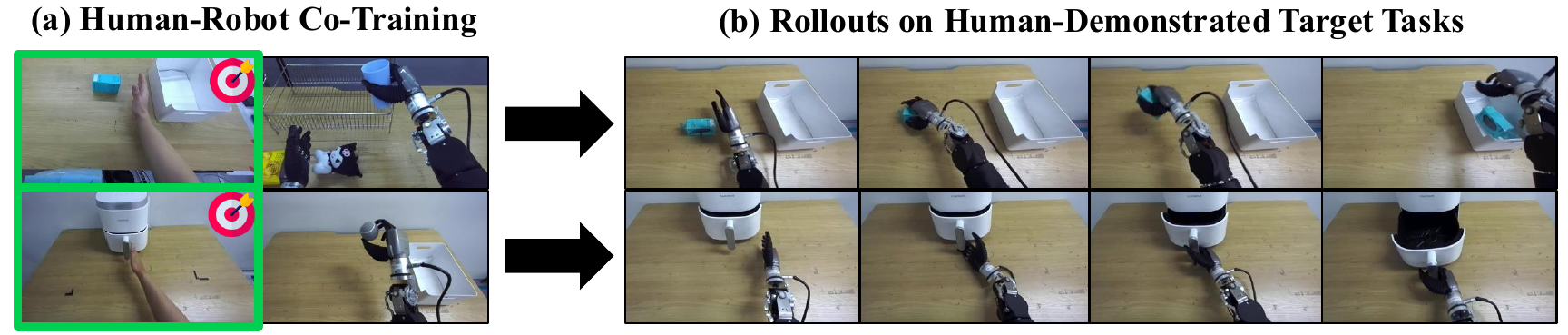}
    \caption{
\textbf{Qualitative human-to-robot transfer rollouts.}
(a) \method{} is trained jointly on human and robot demonstrations.
(b) The trained policy executes target tasks introduced through human demonstrations, including object transfer on box, followed by skill transfer on opening.
}
    \label{fig:exp_setup}
\end{figure*}

Aligning actions alone does not resolve the human-robot perceptual gap, so we anchor VLM conditioning features to hand geometry shared across embodiments (\Cref{fig:pipeline-b}).

We ground VLM features using the current wrist pose and five wrist-local fingertip positions, denoted by $\mathbf g_t$, obtained from reconstructed human motion or robot configurations.
A shared residual adapter $A_\phi$ and training-only readout $R_\omega$ produce
$\mathbf h_t^{\mathrm{geo}}=\mathbf h_t+A_\phi(\mathbf h_t)$ and
$\widehat{\mathbf g}_t=R_\omega(\operatorname{Pool}(\mathbf h_t^{\mathrm{geo}}))$,
% optimized with a masked Smooth-L1 loss $\mathcal L_{\mathrm{geo}}$, 
where $\operatorname{Pool}$ denotes mean pooling over the VLM token sequence.
We define $\mathcal L_{\mathrm{geo}}$ as the masked Smooth-L1 loss between
$\widehat{\mathbf g}_t$ and $\mathbf g_t$ over valid geometry coordinates.
The action expert is conditioned on $\mathbf h_t^{\mathrm{geo}}$.
Because each observation provides its own geometry target, this alignment requires neither paired human-robot demonstrations nor temporal correspondence.

\vspace{0.01in}
\begin{equation}
    \mathcal L_{\mathrm{VLA}}
    =\mathcal L_{\mathrm{pred}}
    +\lambda_{\mathrm{geo}}\mathcal L_{\mathrm{geo}}.
    \label{eq-vlaloss}
\end{equation}
where $\lambda_{\mathrm{geo}}$ weights the geometry objective. Implementation and optimization details are provided in Appendix~\ref{app:policy_optimization}.
The VLM backbone, tokenizer, and embodiment-specific decoders remain frozen during this stage. 
At deployment, numerical integration of the learned velocity field
$\mathbf v_\theta$ induces the latent-action policy $\pi_\theta$, conditioned on
the robot observation, proprioceptive state, and instruction:
\begin{equation}
    \mathbf Z_t^{\mathrm{gen}}
    \sim
    \pi_\theta(\cdot\mid\mathbf h_t^{\mathrm{geo}},\mathbf s_t),
    \qquad
    \mathbf A_t^{\mathrm{cmd}}
    =
    D_H(\mathbf Z_t^{\mathrm{gen}})
    \label{eq:deploy}
\end{equation}
The decoder output is converted to robot commands using its prescribed inverse normalization and coordinate mapping. Both decoders reconstruct the same robot-compatible wrist-hand representation. Although deployment uses $D_H$, the shared encoder and latent-action policy are learned from both human and robot demonstrations.

\section{Experiments}
\label{sec:experiments}

We evaluate \method{} in two stages. First, under controlled task-specific fine-tuning, we test whether human demonstrations transfer target objects and target skills to a robot beyond vanilla human-robot co-training. Second, we test whether the same cross-embodiment principle remains useful when incorporated into heterogeneous multi-embodiment training.

\subsection{Experimental setup}
\label{sec:implementation}

Our real-world experiments use an OpenArm humanoid with Inspire hands, with all policies initialized from the robot-pretrained RLDX-1-IMG~\citep{kim2026rldx}; human demonstrations use the wrist-hand representation in~\Cref{sec:shared-action}.
We compare matched co-training with separate embodiment-specific interfaces against HaWoR and Ego-Pi-based shared-action baselines, all without our geometry alignment or latent tokenizer and using the same adaptation data and training budget (see~\Cref{tab:data_conditions,tab:appendix_baselines}).

We study two axes of human-to-robot transfer.
\emph{Object transfer} introduces target objects through human pick-and-place demonstrations while retaining prior robot pick-and-place experience.
\emph{Skill transfer} introduces opening and closing interactions absent from the source robot demonstrations (see~\Cref{fig:exp_setup}).
For these OpenArm tasks, we assign weighted task scores (0-100) based on execution progress, as detailed in Appendix~\ref{app:evaluation_protocol}.

\begin{figure*}[t]
    \centering
    \captionsetup{font=small, skip=3pt}

    % ==============================
    % Row 1: RB-Y1 / HT
    % ==============================
    \begin{minipage}[t]{0.52\textwidth}
        \centering
        \vspace{0pt}
        \includegraphics[width=\linewidth]{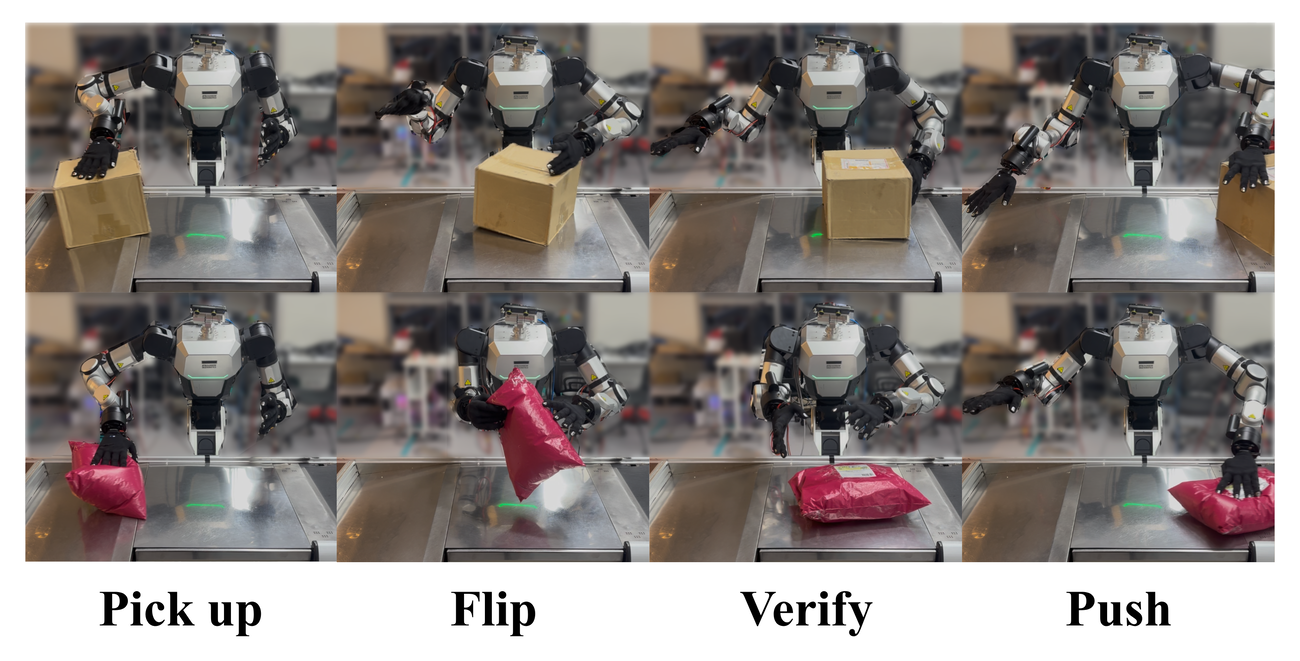}
        \captionof{figure}{
            \textbf{RB-Y1 parcel-handling task.}
            Box flipping (top) and polybag handling (bottom).
        }
        \label{fig:rby1}
    \end{minipage}
    \hfill
    \begin{minipage}[t]{0.44\textwidth}
        \centering
        \vspace{0pt}
        \includegraphics[width=\linewidth]{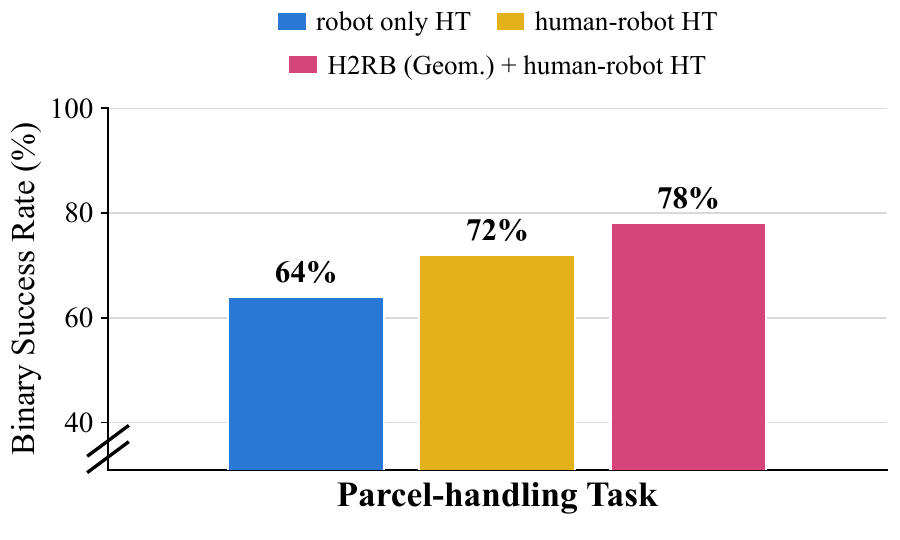}
        \captionof{figure}{
            \textbf{Human-robot HT.}
            Human-robot HT and geometry grounding improves robot-only downstream success.
        }
        \label{fig:ht}
    \end{minipage}

    % ==============================
    % Row 2: Transfer / Instruction
    % ==============================
    \begin{minipage}[t]{0.52\textwidth}
        \centering
        \vspace{0pt}
        \includegraphics[width=\linewidth]{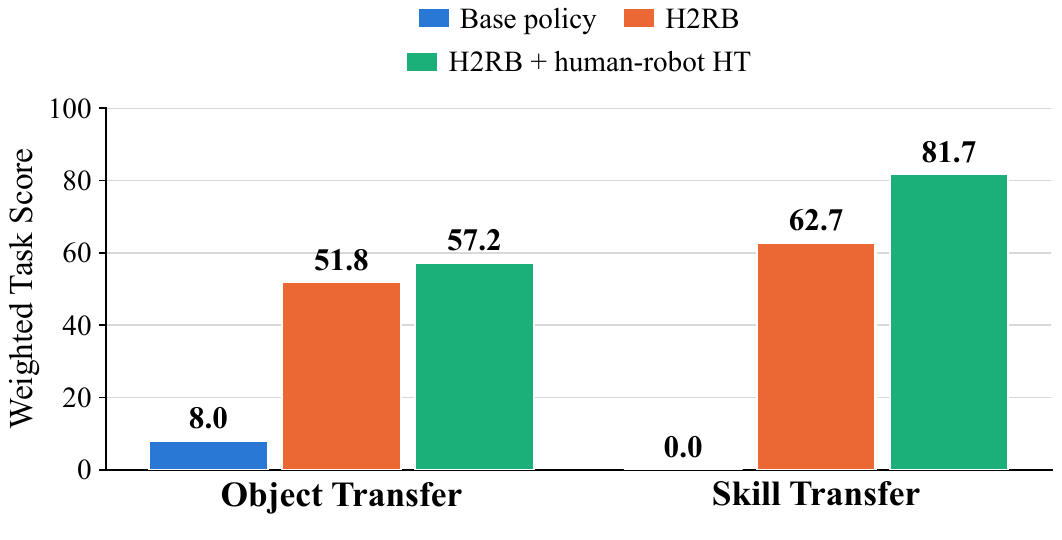}
        \captionof{figure}{
            \textbf{Effect of human-robot HT.}
            HT improves transfer abilities, showing the scalability of \method{}.
        }
        \label{fig:mid-training}
    \end{minipage}
    \hfill
    \begin{minipage}[t]{0.44\textwidth}
        \centering
        \vspace{0pt}
        \includegraphics[width=\linewidth]{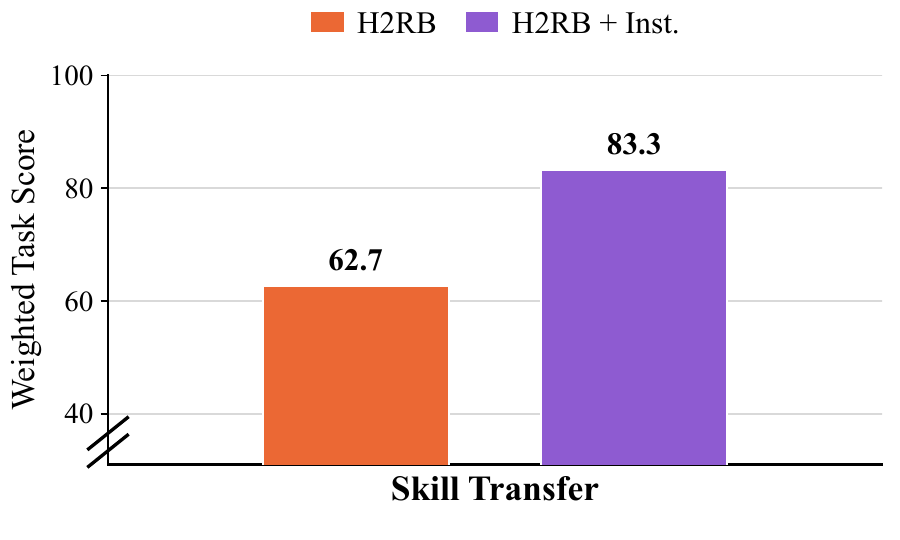}
        \captionof{figure}{
            \textbf{Instruction conditioning.}
            Instruction conditioning improves skill transfer abilities.
        }
        \label{fig:instruction-conditioning}
    \end{minipage}

\end{figure*}

For heterogeneous training evaluation, we additionally consider an RB-Y1 parcel-handling task, where the robot flips boxes and polybags to expose their shipping labels before pushing them onward (see \Cref{fig:rby1}). RB-Y1 uses robot-only fine-tuning and binary success over 50 trials.

% ------------------------
% I: Object transfer
% ------------------------
\begin{table}[t]
\centering
\small
\caption{
\textbf{Human-to-robot object transfer.}
Robot in-domain reports performance on a robot-data-trained task, while
Human transfer evaluates target tasks introduced through human demonstrations with 20 ID and 10 OOD rollouts per target object. Robot in-domain (Doll) is evaluated separately with 20 rollouts.
Each entry is the weighted task score (0-100).
}

\label{tab:main_object_transfer}

\setlength{\tabcolsep}{3pt}
\renewcommand{\arraystretch}{1.08}

\begin{adjustbox}{max width=\linewidth}
\begin{tabular}{lcccccccccc}
    \toprule
    & \multicolumn{1}{c}{Robot in-domain}
    & \multicolumn{9}{c}{Human transfer} \\
    \cmidrule(lr){2-2}
    \cmidrule(lr){3-11}

    & \multirow{2}{*}{Doll}
    & \multicolumn{2}{c}{Box}
    & \multicolumn{2}{c}{Ball}
    & \multicolumn{2}{c}{Towel}
    & \multicolumn{3}{c}{Avg.} \\
    \cmidrule(lr){3-4}
    \cmidrule(lr){5-6}
    \cmidrule(lr){7-8}
    \cmidrule(lr){9-11}

    Method &
    & ID & OOD
    & ID & OOD
    & ID & OOD
    & ID & OOD & All \\
    \midrule

    Base policy human-only
    & 31.5
    & 41.0 & 21.0
    & 31.5 & 15.0
    & 28.5 & 24.0
    & 33.7 & 20.0 & 29.1 \\

    Base policy in-domain-only
    & 55.0
    & 22.5 & 9.0
    & 16.5 & 9.0
    & 3.0 & 6.0
    & 14.0 & 8.0 & 12.0 \\

    \midrule

    Base policy co-train
    & 18.0
    & 13.5 & 9.0
    & 3.0 & 6.0
    & 10.5 & 3.0
    & 9.0 & 6.0 & 8.0 \\

    + HaWoR-shared action~\citep{zhang2025hawor}
    & 10.5
    & 10.5 & 12.0
    & 10.5 & 15.0
    & 9.0 & 9.0
    & 10.0 & 12.0 & 10.7 \\

    + Ego-Pi-shared action~\citep{kim2026ego}
    & 30.0
    & 45.0 & 25.0
    & 18.5 & 15.0
    & 27.0 & 24.0
    & 30.2 & 21.3 & 27.2 \\

    \rowcolor{gray!10}
    \textbf{+ \method{} (Ours)}
    & \textbf{49.0}
    & \textbf{60.0} & \textbf{39.0}
    & \textbf{50.0} & \textbf{47.0}
    & \textbf{55.0} & \textbf{50.0}
    & \textbf{55.0} & \textbf{45.3} & \textbf{51.8} \\

    \bottomrule
\end{tabular}
\end{adjustbox}
\end{table}

% ------------------------
% II: Skill transfer
% ------------------------
\begin{table}[t]
\centering
\small
\caption{
\textbf{Human-to-robot skill transfer.}
Robot in-domain reports performance on a robot-data-trained task, while
Human transfer evaluates target tasks introduced through human demonstrations.
Each entry is the weighted task score (0-100).
}
\label{tab:main_skill_transfer}

\renewcommand{\arraystretch}{1.08}

\begin{tabular}{l c c c c}
    \toprule
    & \multicolumn{1}{c}{Robot in-domain}
    & \multicolumn{3}{c}{Human transfer} \\
    \cmidrule(lr){2-2}
    \cmidrule(lr){3-5}
    Method
    &
    \makecell{Box PnP}
    &
    \makecell{Open}
    &
    \makecell{Close}
    &
    \makecell{Avg.}
    \\

    \midrule

    Base policy human-only
    & 0.0
    & {36.0}
    & {22.0}
    & {29.0} \\

    Base policy in-domain-only
    & 80.5
    & 0.0
    & 0.0
    & 0.0 \\
    \midrule
    Base policy co-train
    & 16.5
    & 0.0
    & 0.0
    & 0.0 \\

    + HaWoR-shared action~\citep{zhang2025hawor}
    & 19.5
    & 0.0
    & 0.0
    & 0.0 \\

    + Ego-Pi-shared action~\citep{kim2026ego}
    & 70.5
    & 0.0
    & 0.0
    & 0.0 \\

    \rowcolor{gray!10}
    \textbf{+ \method{} (Ours)}
    & \textbf{81.0}
    & \textbf{50.7}
    & \textbf{74.7}
    & \textbf{62.7} \\

    \bottomrule
\end{tabular}

\end{table}
\subsection{Human-to-robot task transfer}
\label{sec:task_transfer}

\vspace{0.01in}
\noindent{\bf Object transfer results.}
\Cref{tab:main_object_transfer} summarizes transfer to objects introduced only through human demonstrations.
The robot-only policy performs well on in-domain Doll PnP (55.0) but poorly on unseen objects (12.0), while vanilla co-training further degrades transfer to 8.0.
In contrast, \method{} improves object transfer to 51.8 (+43.8) while retaining strong in-domain performance (49.0).
Under object-placement shift, \method{} also outperforms Ego-Pi-shared action in both ID (30.2 to 55.0) and OOD (21.3 to 45.3) settings.
These gains suggest that bridging the human-robot perceptual gap enables robust transfer to novel objects and locations while retaining robot-pretrained capabilities.
Further breakdowns and geometry-grounding ablations are provided in Appendix~\ref{app:additional_transfer}.

\vspace{0.01in}
\noindent{\bf Skill transfer results.}
Skill transfer exposes a different bottleneck, as shown in \Cref{tab:main_skill_transfer}. 
HaWoR-shared action fails to transfer Open/Close and degrades in-domain performance to 19.5, while Ego-Pi recovers in-domain execution to 70.5 but still achieves 0.0 on both target skills. This suggests that a shared executable action space alone is insufficient for skill transfer; \method{} instead transfers both skills while further improving in-domain performance from 0.0 to 81.0.
This result highlights the importance of separating shared latent skill representation from embodiment-specific action realization when transferring behaviors across different motion distributions.
Representative successful rollouts for each transfer task are shown in \Cref{fig:qualitative-success}.

\subsection{Scaling to heterogeneous human-robot data}
\label{sec:human_scaling}

We further adapt the pretrained VLA backbone through large-scale heterogeneous human-robot training (HT) across multiple embodiments, followed by downstream task-specific fine-tuning (FT). We evaluate its benefits in two downstream settings: human-to-robot transfer on OpenArm and robot-only adaptation on RB-Y1. Training details are provided in Appendix~\ref{app:midtraining}.

For OpenArm, heterogeneous training improves object transfer from 51.8 to 57.2 (+5.4) and skill transfer from 62.7 to 81.7 (+19.0), as shown in~\Cref{fig:mid-training}. These gains demonstrate the benefit of additional heterogeneous training beyond task-specific co-training. For RB-Y1, parcel-orientation success improves from 64\% with robot-only training to 72\% with human-robot HT and 78\% with geometry-grounded HT, as shown in~\Cref{fig:ht}. All variants use robot-only downstream FT without the shared action tokenizer, demonstrating the benefit of geometry grounding beyond human-to-robot task transfer. Together with the OpenArm results, it suggests that learning compatible action-relevant representations benefits downstream adaptation across different robots and task settings.

\subsection{Enhancing skill transfer with instruction conditioning}
\label{sec:component_analysis}

Separately from heterogeneous training, we examine whether stronger language conditioning can improve the transferred behavior for skill transfer.
We treat the instruction as an explicit skill signal by concatenating its representation with the state-conditioned vision-language latent and re-injecting it throughout the expert layers.
As shown in ~\Cref{fig:instruction-conditioning}, this improves skill transfer from 62.7 to 83.3 (+20.6), suggesting that sustained instruction access helps preserve task semantics during cross-embodiment execution.
Implementation details are provided in Appendix~\ref{app:instruction_implementation}.
\section{Analysis}
\label{sec:analysis}

Human-to-robot transfer requires incorporating human-provided task structure while aligning cross-embodiment representations without discarding robot-specific execution dynamics.
We analyze this separation through transfer failures, execution speed, and the resulting human-robot representation.

\subsection{Qualitative analysis of transfer failures}
\label{sec:source_analysis}

\begin{figure*}[t]
    \centering
    \captionsetup[subfigure]{skip=1mm}

    % (a)
    \begin{subfigure}[t]{0.48\textwidth}
        \centering
        \includegraphics[width=\linewidth]{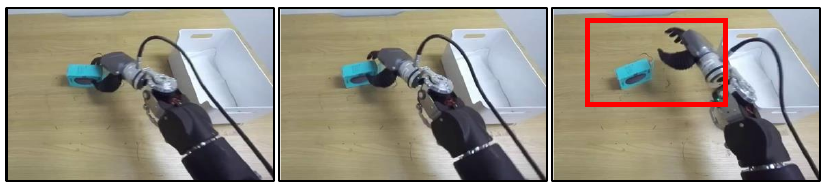}
        \caption{Human-only policy : Grasping failure}
        \label{fig:qualitative-failures-a}
    \end{subfigure}
    \hfill
    % (b)
    \begin{subfigure}[t]{0.48\textwidth}
        \centering
        \includegraphics[width=\linewidth]{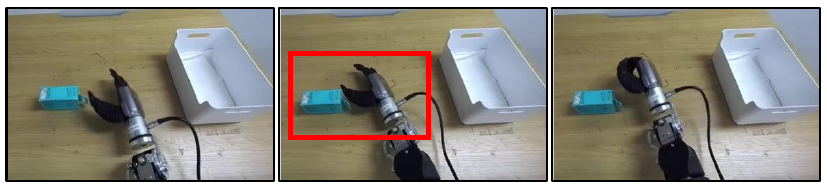}
        \caption{Robot-only policy : Target-location mismatch}
        \label{fig:qualitative-failures-b}
    \end{subfigure}

    \vspace{1mm}

    % (c)
    \begin{subfigure}[t]{0.48\textwidth}
        \centering
        \includegraphics[width=\linewidth]{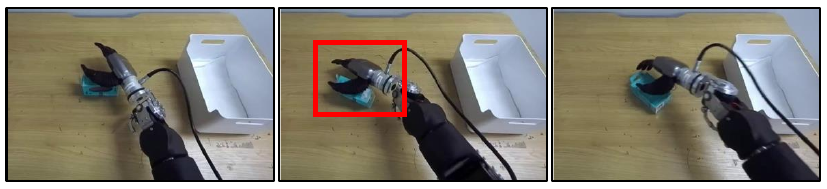}
        \caption{Ego-Pi-shared action : Invalid grasp pose}
        \label{fig:qualitative-failures-c}
    \end{subfigure}
    \hfill
    % (d)
    \begin{subfigure}[t]{0.48\textwidth}
        \centering
        \includegraphics[width=\linewidth]{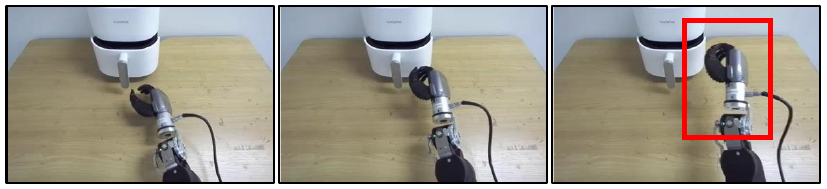}
        \caption{Ego-Pi-shared action : Task-motion mismatch}
        \label{fig:qualitative-failures-d}
    \end{subfigure}

    \caption{
    \textbf{Failure case analysis of human-to-robot transfer.}
    (a) Fails to complete the grasp.
    (b) Unrelated to the target object location.
    (c) Fails to produce a hand pose suitable for grasping.
    (d) Executes a robot-observed pick-and-place motion in human-observed task environment.
    }
    \label{fig:qualitative-failures}
\end{figure*}

Human-only and robot-only policies exhibit complementary failure modes.
Human-only policies capture the demonstrated interaction pattern but fail near contact due to inaccurate wrist alignment and grasp configuration (\Cref{fig:qualitative-failures-a}), whereas robot-only policies produce executable robot motions but fail to localize or reproduce behaviors absent from robot demonstrations (\Cref{fig:qualitative-failures-b}).
Thus, human data provide the target behavior, while robot data provide embodiment-specific knowledge.

Combining the two sources in a shared action format is still not a fundamental solution.
The shared-action baseline produces invalid grasp poses, indicating that coordinate-level alignment does not resolve the remaining perceptual mismatch (\Cref{fig:qualitative-failures-c}).
Moreover, robot-conditioned predictions can remain closer to the robot latent distribution and revert to familiar robot behaviors even in human task deployment, such as pick-and-place instead of the human-demonstrated opening motion (\Cref{fig:qualitative-failures-d}).
These failures motivate bridging the two domains beyond a common physical action format. Accordingly, \method{} reduces the latent-level human-robot gap while retaining embodiment-specific decoders for source-appropriate action realization.

\subsection{Cross-embodiment gap analysis}
\label{sec:latent-analysis}

\begin{table*}[t]
    \centering
    \captionsetup[subtable]{skip=0pt}

    \caption{
    \textbf{Analysis of the decoupled-decoder latent tokenizer.}
    (a) Effect of the action tokenizer and cross-domain latent regression on
    object transfer, skill transfer, and normalized human-robot latent gap.
    (b) Motion speed and average success of embodiment-specific reconstruction decoders.
    }
    \label{tab:action-tokenizer-analysis}

    \begin{subtable}[t]{0.52\textwidth}
        \centering
        \small

        \caption{Action tokenizer ablation}
        \label{tab:action-tokenizer-ablation}
        \vspace{0.06in}
        \resizebox{\linewidth}{!}{%
        \begin{tabular}{c c c c c}
            \toprule
            Action
            & Cross-domain
            & Object
            & Skill
            & Human-robot \\
            Tokenizer
            & latent regularization
            & transfer
            & transfer
            & latent gap \\
            \midrule
            \textcolor{red}{\xmark}
            & \textcolor{red}{\xmark}
            & 34.7
            & 0.0
            & - \\
            \textcolor{green!60!black}{\cmark}
            & \textcolor{red}{\xmark}
            & 43.3
            & 49.7
            & 3.81 \\
            \textcolor{green!60!black}{\cmark}
            & \textcolor{green!60!black}{\cmark}
            & \textbf{51.8}
            & \textbf{62.7}
            & \textbf{1.53} \\
            \bottomrule
        \end{tabular}%
        }\par

    \end{subtable}%
    \hfill%
    \begin{subtable}[t]{0.45\textwidth}
        \centering
        \small

        \caption{Embodiment-specific decoding}
        \label{tab:action-tokenizer-decoding}
        \vspace{0.06in}
        \resizebox{\linewidth}{!}{%
        \begin{tabular}{l c c}
            \toprule
            Training source
            & Motion speed
            & Weighted \\
            & (cm/s)
            & task score \\
            \midrule
            Robot demonstrations
            & 2.52
            & \textbf{65.3} \\
            Human demonstrations
            & \textbf{7.36} ($\times 2.92$)
            & \underline{62.7} \\
            \bottomrule
        \end{tabular}%
        }\par

    \end{subtable}%
\end{table*}

% -----------------
% Failure analysis rollouts
% -----------------

\begin{figure*}[t]
    \centering
    ~~
    % (a) Conditioning-space adaptation
    \begin{subfigure}[t]{0.9\textwidth}
        \centering
        \begin{minipage}[t]{0.47\linewidth}
            \centering
            \includegraphics[width=\linewidth,trim=8 8 8 8,clip]
            {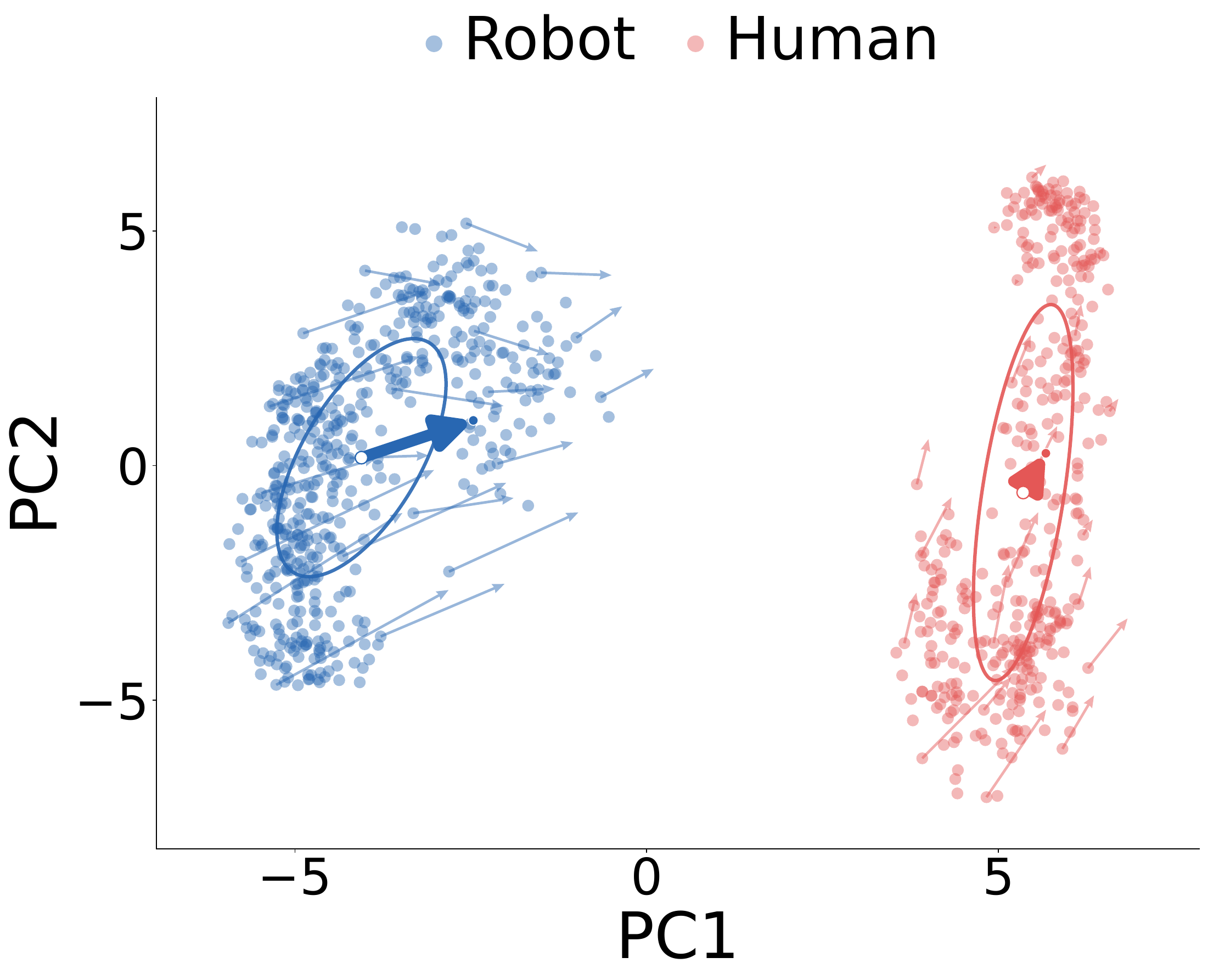}
        \end{minipage}\hfill
        \begin{minipage}[t]{0.47\linewidth}
            \centering
            \includegraphics[width=\linewidth,trim=8 8 8 8,clip]
            {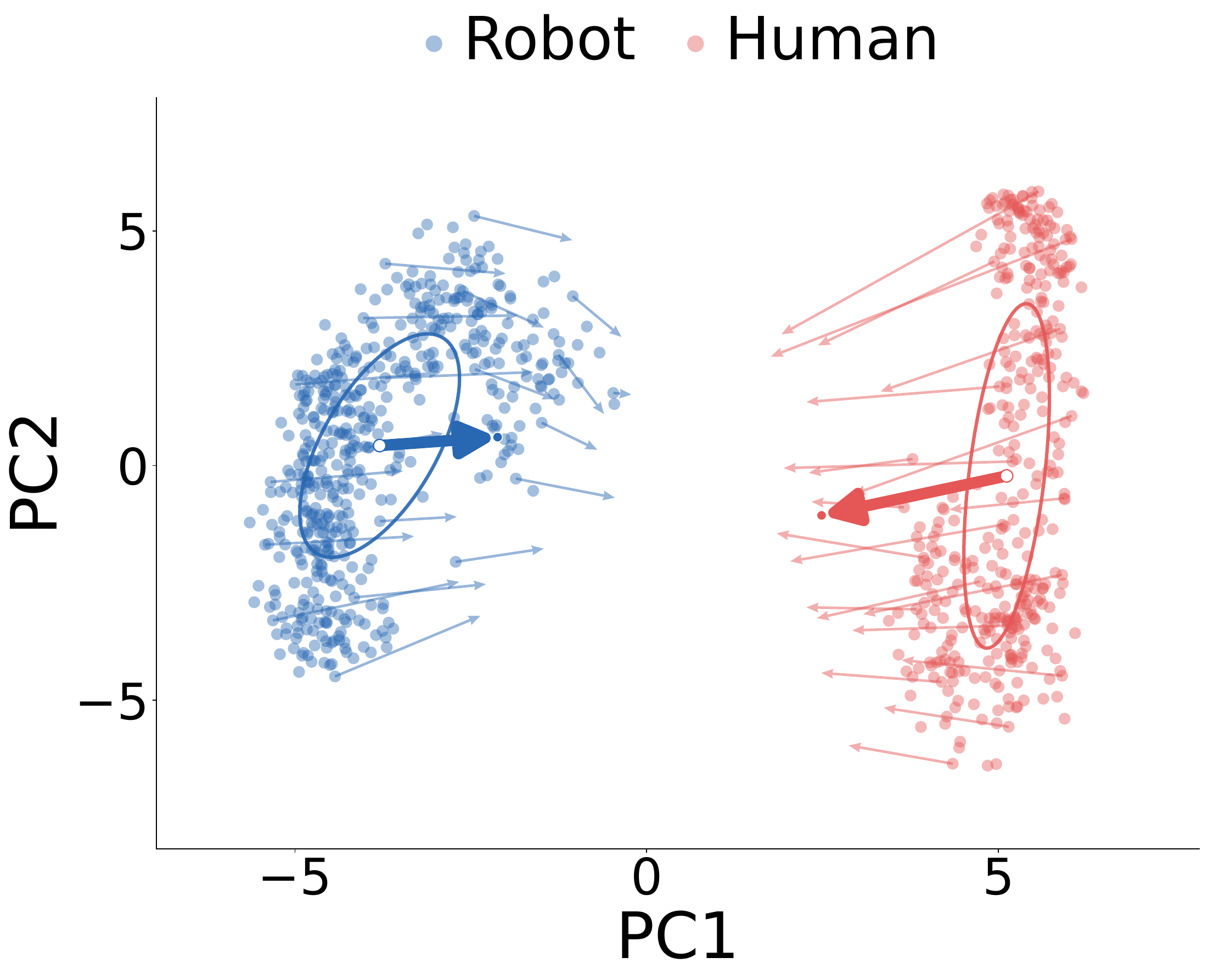}
        \end{minipage}

        \vspace{1mm}
        \caption{Conditioning-space adaptation}
        \label{fig:latent-analysis-conditioning}
    \end{subfigure}
    \vspace{3mm}

    % (b) Action-latent adaptation
    \begin{subfigure}[t]{0.9\textwidth}
        \centering
        \begin{minipage}[t]{0.47\linewidth}
            \centering
            \includegraphics[width=\linewidth,trim=8 8 8 8,clip]
            {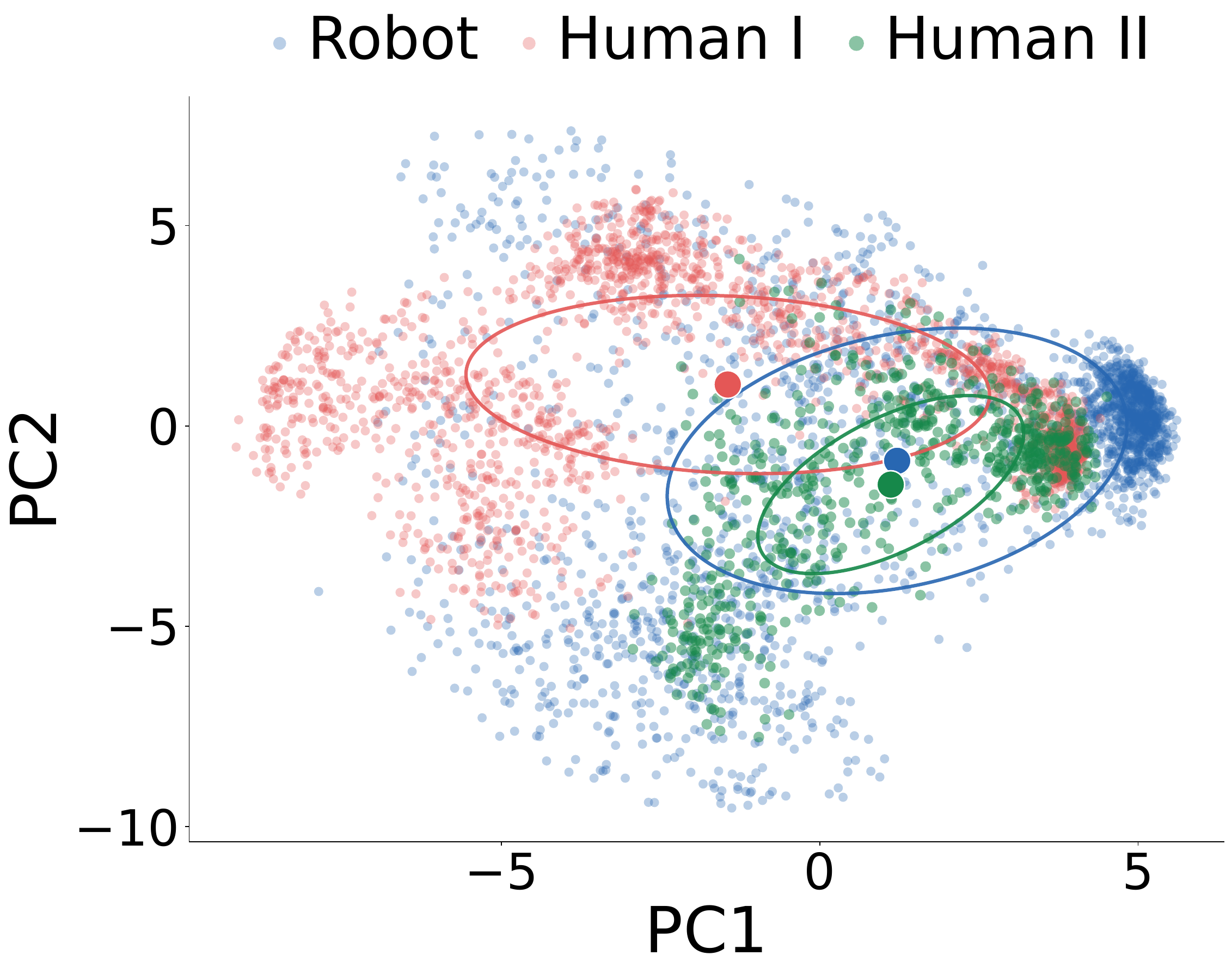}
        \end{minipage}\hfill
        \begin{minipage}[t]{0.47\linewidth}
            \centering
            \includegraphics[width=\linewidth,trim=8 8 8 8,clip]
            {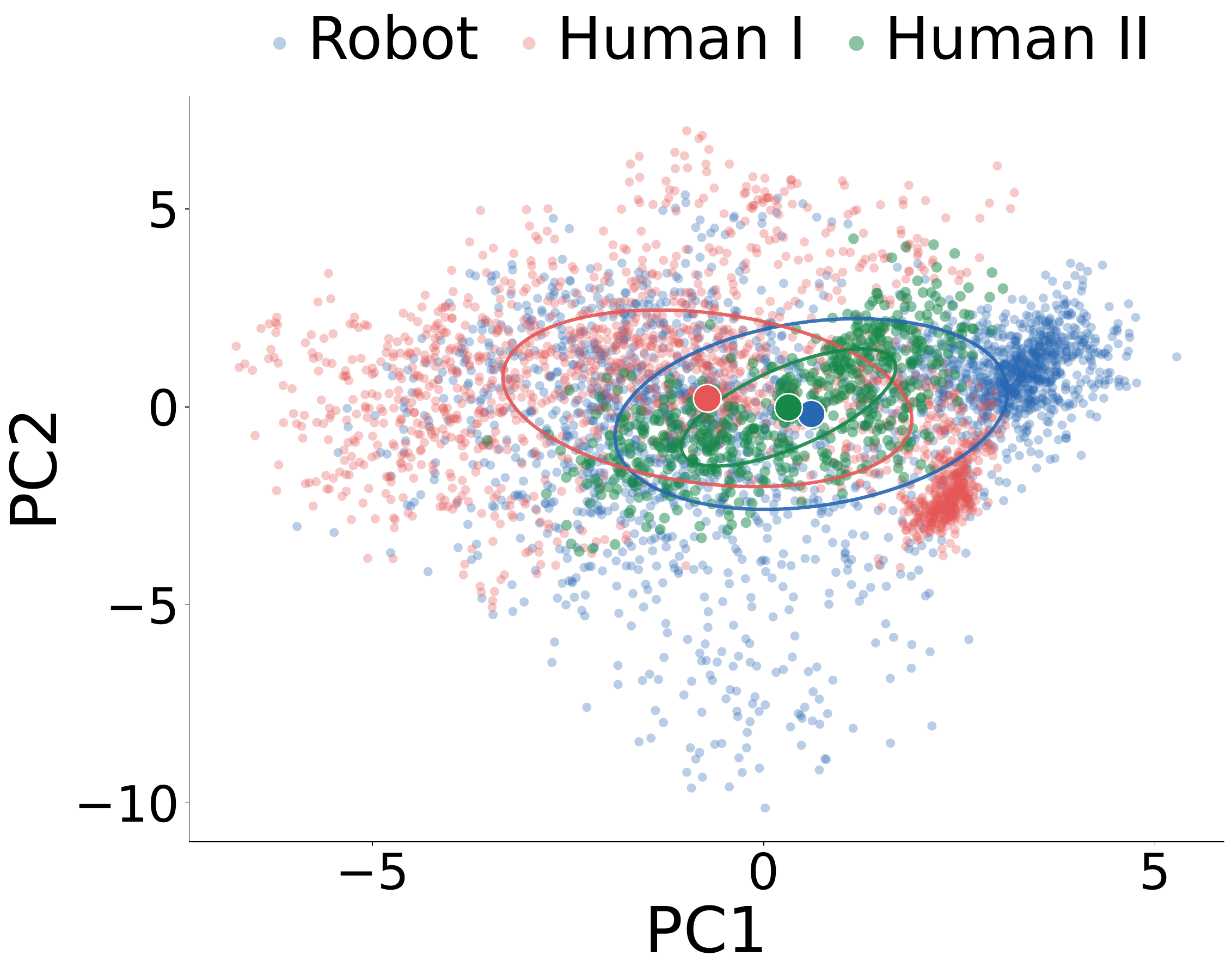}
        \end{minipage}

        \vspace{1mm}
        \caption{Action-latent adaptation}
        \label{fig:latent-analysis-action}
    \end{subfigure}

    \caption{
    \textbf{PCA visualization of cross-embodiment representation adaptation.}
        (a) Vision-language conditioning features with hand geometry-anchored alignment alone
        (left) and \method{} (right).
        Arrows show adapter-induced displacements for the same observation.
        (b) Action-tokenizer latents without (left) and with (right)
        cross-domain latent regression, using per-tokenizer PCA.
        Human I and II denote human tasks trained in within-skill object transfer
and skill transfer, respectively.
    }
    \label{fig:latent-analysis}

    \vspace{-10pt}
\end{figure*}

\vspace{0.01in}
\noindent{\bf Hand geometry-anchored alignment reduces the conditioning-space gap.}
Hand geometry-anchored alignment supervises human and robot observations through a common hand-geometry target without requiring paired human-robot data.
Compared with the Ego-Pi-shared action baseline, the PCA projection in \Cref{fig:latent-analysis-conditioning} shows that \method{} shifts robot conditioning features toward the human distribution, reducing the human-robot gap.
This suggests that grounding VLM features in shared hand geometry makes the two embodiments more compatible in the conditioning space used for downstream action prediction.

\vspace{0.01in}
\noindent{\bf Decoupled-decoder latent tokenizer progressively aligns the prediction space.}
~\Cref{tab:action-tokenizer-ablation} shows that introducing the action tokenizer improves object transfer from 34.7 to 43.3 and enables skill transfer from 0.0 to 49.7.
Cross-domain regularization further reduces the normalized human-robot latent gap (defined in Appendix~\ref{app:normalized_gap}) from 3.81 to 1.53, a 59.8\% reduction, while improving object and skill transfer to 51.8 and 62.7, respectively.
This alignment is particularly important at deployment, where robot-conditioned predictions can otherwise remain shifted toward the robot latent distribution, even for tasks learned from human data.
The two-dimensional projections in~\Cref{fig:latent-analysis-action} provide a qualitative view of this alignment, showing how cross-domain regression brings the human and robot latent distributions closer in the action latent space.

\vspace{0.01in}
\noindent{\bf Faster task execution from human data.}
Human demonstrations can provide faster task-specific motion while requiring substantially less collection time and robot interaction. To test whether these dynamics are preserved after transfer, we collected 100 robot teleoperation episodes for the same Open/Close tasks, with 50 episodes per task, and trained a robot-only policy. 
As shown in~\Cref{tab:action-tokenizer-decoding}, the human-transferred policy reaches 7.36 cm/s versus 2.52 cm/s for the robot-only policy, with comparable task performance (62.7 vs. 65.3).
Because the embodiment-specific decoders preserve distinct action dynamics, the transferred policy can retain the faster motion from human demonstrations even when deployed on the robot.

\section{Related Work}
\label{sec:related_work}

\vspace{0.01in}
\noindent{\bf Learning from human demonstrations.}
Human videos provide policy supervision through reconstructed or tracked hand motion~\citep{zhang2025hawor,hoque2026egodex,li2025scalable,qin2022dexmv} or latent actions inferred from frame transitions~\citep{ye2025latent,chen2026villa}, supporting co-training and pretraining in works such as EgoMimic, EgoVLA, and EgoScale~\citep{kareer2025egomimic,yang2025egovla,zheng2026egoscale,tao2025dexwild}.
Ego-Pi~\citep{kim2026ego} transfers task semantics through co-training in a common wrist-hand space, while HumanEgo~\citep{wang2026humanego} learns from human demonstrations without robot data using hand-object representations.
We focus on adapting a robot-pretrained VLA to human-demonstrated target tasks while retaining experience from robot data.

\vspace{0.01in}
\noindent{\bf Cross-embodiment representation alignment.}
Human-data pretraining builds cross-embodiment priors~\citep{yang2025egovla,zheng2026egoscale}, while explicit alignment relies on paired demonstrations~\citep{xie2026human2robot,zhu2026harp} or trajectory-based pseudo-pairs~\citep{punamiya2025egobridge}.
PHANTOM and Masquerade instead synthesize robot-domain observations through inpainting and rendering~\citep{lepert2025phantom,lepert2025masquerade}.
\method{} grounds pretrained VLA features in hand geometry independently for each observation, providing a shared physical reference without paired data, trajectory matching, image synthesis, or prior human-robot backbone training.

\vspace{0.01in}
\noindent{\bf Latent action representations and decoding.}
Diffusion and flow-based policies generate continuous actions~\citep{chi2025diffusion, black2024pi_0,
ze2024dp3,zhang2025flowpolicy}, while latent diffusion separates generation from output reconstruction~\citep{rombach2022high, peebles2023scalable}.
LAD aligns retargeted action pairs~\citep{bauer2025latent}, and XL-VLA learns cross-hand latent representations~\citep{jiang2026cross}, both retaining embodiment-specific decoders for heterogeneous end-effectors.
H2RB shares action latents across human-robot demonstrations while preserving source-specific motion dynamics through separate temporal decoders.
\section{Conclusion}
We present \method{}, a framework for adapting pretrained VLAs from unpaired human and robot demonstrations.
Rather than relying on implicit cross-embodiment alignment, \method{} bridges perception through hand geometry-anchored alignment and action through a shared latent space with embodiment-specific decoders.
Real-world experiments show improved object and skill transfer, further gains from heterogeneous training, and benefits for robot-only adaptation on a different embodiment.
These findings highlight the importance of learning reusable action-relevant structure from heterogeneous experience without enforcing identical action dynamics.
We hope our approach helps bridge scalable human demonstration data and robot task adaptation, enabling pretrained VLAs to acquire new behaviors with human demonstration data.

\clearpage

\section*{Acknowledgements}
This research was supported by the “Advanced GPU Utilization Support Program” funded by the Government of the Republic of Korea (Ministry of Science and ICT).
We thank the members of Algorithmic General Intelligence Laboratory (AGI-LAB) of KAIST, for helpful support in conducting real-world experiments.

\bibliographystyle{plainnat}
\bibliography{paper}

@String(CVPR  = {IEEE Conference on Computer Vision and Pattern Recognition})

@String(ICCV = {IEEE International Conference on Computer Vision})

@String(ICRA = {IEEE International Conference on Robotics and Automation})

@article{kim2026rldx,
  title={Rldx-1 technical report},
  author={Kim, Dongyoung and Jang, Huiwon and Koo, Myungkyu and Jang, Suhyeok and Kim, Taeyoung and Kim, Beomjun and Yoon, Byungjun and Jang, Changsung and Choi, Daewon and Han, Dongsu and others},
  journal={arXiv preprint arXiv:2605.03269},
  year={2026}
}

@inproceedings{kim2026ego,
  title={Ego-pi: Vla fine-tuning for ego-centric human and robot data},
  author={Kim, Ji Woong and Wang, Ke and Fu, Zipeng and Chen, Sirui and Lai, Jeff and Finn, Chelsea and others},
  booktitle={Proceedings of the IEEE/CVF Conference on Computer Vision and Pattern Recognition},
  pages={1515--1524},
  year={2026}
}

@article{lipman2022flow,
  title={Flow matching for generative modeling},
  author={Lipman, Yaron and Chen, Ricky TQ and Ben-Hamu, Heli and Nickel, Maximilian and Le, Matt},
  journal={arXiv preprint arXiv:2210.02747},
  year={2022}
}

@inproceedings{zhang2025hawor,
  title={Hawor: World-space hand motion reconstruction from egocentric videos},
  author={Zhang, Jinglei and Deng, Jiankang and Ma, Chao and Potamias, Rolandos Alexandros},
  booktitle={2025 IEEE/CVF Conference on Computer Vision and Pattern Recognition (CVPR)},
  pages={1805--1815},
  year={2025},
  organization={IEEE}
}

@inproceedings{hoque2026egodex,
  title={Egodex: Learning dexterous manipulation from large-scale egocentric video},
  author={Hoque, Ryan and Huang, Peide and Yoon, David and Zhang, Jian and others},
  booktitle={International Conference on Learning Representations},
  volume={2026},
  pages={4218--4237},
  year={2026}
}

@article{li2025scalable,
  title={Scalable vision-language-action model pretraining for robotic manipulation with real-life human activity videos},
  author={Li, Qixiu and Deng, Yu and Liang, Yaobo and Luo, Lin and Zhou, Lei and Yao, Chengtang and Zeng, Lingqi and Feng, Zhiyuan and Liang, Huizhi and Xu, Sicheng and others},
  journal={arXiv preprint arXiv:2510.21571},
  year={2025}
}

@article{wang2026humanego,
  title={HumanEgo: Zero-Shot Robot Learning from Minutes of Human Egocentric Videos},
  author={Wang, Zhi and He, Botao and Yu, Kelin and Lee, Seungjae and Gao, Ruohan and Huang, Furong and Aloimonos, Yiannis},
  journal={arXiv preprint arXiv:2605.24934},
  year={2026}
}

@inproceedings{punamiya2025egobridge,
  title={Egobridge: Domain adaptation for generalizable imitation from egocentric human data},
  author={Punamiya, Ryan and Patel, Dhruv and Aphiwetsa, Patcharapong and Kuppili, Pranav and Zhu, Lawrence Y and Kareer, Simar and Hoffman, Judy and Xu, Danfei},
  booktitle={Human to Robot: Workshop on Sensorizing, Modeling, and Learning from Humans},
  year={2025}
}

@article{zhu2026harp,
  title={HARP-VLA: Human-Robot Aligned Representation Learning for Vision-Language-Action Model},
  author={Zhu, Xiang and Yuan, Puzhen and Liu, Yichen and Chen, Jianyu},
  journal={arXiv preprint arXiv:2605.31234},
  year={2026}
}

@article{kim2024openvla,
  title={Openvla: An open-source vision-language-action model},
  author={Kim, Moo Jin and Pertsch, Karl and Karamcheti, Siddharth and Xiao, Ted and Balakrishna, Ashwin and Nair, Suraj and Rafailov, Rafael and Foster, Ethan and Lam, Grace and Sanketi, Pannag and others},
  journal={arXiv preprint arXiv:2406.09246},
  year={2024}
}

@article{bjorck2025gr00t,
  title={Gr00t n1: An open foundation model for generalist humanoid robots},
  author={Bjorck, Johan and Casta{\~n}eda, Fernando and Cherniadev, Nikita and Da, Xingye and Ding, Runyu and Fan, Linxi and Fang, Yu and Fox, Dieter and Hu, Fengyuan and Huang, Spencer and others},
  journal={arXiv preprint arXiv:2503.14734},
  year={2025}
}

@article{black2024pi_0,
  title={$pi\_0$: A Vision-Language-Action Flow Model for General Robot Control},
  author={Black, Kevin and Brown, Noah and Driess, Danny and Esmail, Adnan and Equi, Michael and Finn, Chelsea and Fusai, Niccolo and Groom, Lachy and Hausman, Karol and Ichter, Brian and others},
  journal={arXiv preprint arXiv:2410.24164},
  year={2024}
}

@article{khazatsky2024droid,
  title={Droid: A large-scale in-the-wild robot manipulation dataset},
  author={Khazatsky, Alexander and Pertsch, Karl and Nair, Suraj and Balakrishna, Ashwin and Dasari, Sudeep and Karamcheti, Siddharth and Nasiriany, Soroush and Srirama, Mohan Kumar and Chen, Lawrence Yunliang and Ellis, Kirsty and others},
  journal={arXiv preprint arXiv:2403.12945},
  year={2024}
}

@article{brohan2022rt,
  title={Rt-1: Robotics transformer for real-world control at scale},
  author={Brohan, Anthony and Brown, Noah and Carbajal, Justice and Chebotar, Yevgen and Dabis, Joseph and Finn, Chelsea and Gopalakrishnan, Keerthana and Hausman, Karol and Herzog, Alex and Hsu, Jasmine and others},
  journal={arXiv preprint arXiv:2212.06817},
  year={2022}
}

@article{qiu2025humanoid,
  title={Humanoid policy\~{} human policy},
  author={Qiu, Ri-Zhao and Yang, Shiqi and Cheng, Xuxin and Chawla, Chaitanya and Li, Jialong and He, Tairan and Yan, Ge and Yoon, David J and Hoque, Ryan and Paulsen, Lars and others},
  journal={arXiv preprint arXiv:2503.13441},
  year={2025}
}

@article{yang2025egovla,
  title={Egovla: Learning vision-language-action models from egocentric human videos},
  author={Yang, Ruihan and Yu, Qinxi and Wu, Yecheng and Yan, Rui and Li, Borui and Cheng, An-Chieh and Zou, Xueyan and Fang, Yunhao and Cheng, Xuxin and Qiu, Ri-Zhao and others},
  journal={arXiv preprint arXiv:2507.12440},
  year={2025}
}

@article{chi2024universal,
  title={Universal manipulation interface: In-the-wild robot teaching without in-the-wild robots},
  author={Chi, Cheng and Xu, Zhenjia and Pan, Chuer and Cousineau, Eric and Burchfiel, Benjamin and Feng, Siyuan and Tedrake, Russ and Song, Shuran},
  journal={arXiv preprint arXiv:2402.10329},
  year={2024}
}

@article{bauer2025latent,
  title={Latent action diffusion for cross-embodiment manipulation},
  author={Bauer, Erik and Nava, Elvis and Katzschmann, Robert K},
  journal={arXiv preprint arXiv:2506.14608},
  year={2025}
}

@article{zheng2026egoscale,
  title={Egoscale: Scaling dexterous manipulation with diverse egocentric human data},
  author={Zheng, Ruijie and Niu, Dantong and Xie, Yuqi and Wang, Jing and Xu, Mengda and Jiang, Yunfan and Casta{\~n}eda, Fernando and Hu, Fengyuan and Tan, You Liang and Fu, Letian and others},
  journal={arXiv preprint arXiv:2602.16710},
  year={2026}
}

@inproceedings{ye2025latent,
  title={Latent action pretraining from videos},
  author={Ye, Seonghyeon and Jang, Joel and Jeon, Byeongguk and Joo, Se June and Yang, Jianwei and Peng, Baolin and Mandlekar, Ajay and Tan, Reuben and Chao, Yu-Wei and Lin, Bill Yuchen and others},
  booktitle={International Conference on Learning Representations},
  volume={2025},
  pages={28213--28239},
  year={2025}
}

@inproceedings{chen2026villa,
  title={Villa-x: enhancing latent action modeling in vision-language-action models},
  author={Chen, Xiaoyu and Wei, Hangxing and Zhang, Pushi and Zhang, Chuheng and Wang, Kaixin and Guo, Yanjiang and Yang, Rushuai and Wang, Yucen and Xiao, Xinquan and Zhao, Li and others},
  booktitle={International Conference on Learning Representations},
  volume={2026},
  pages={70673--70703},
  year={2026}
}

@article{dai2026conla,
  title={Conla: Contrastive latent action learning from human videos for robotic manipulation},
  author={Dai, Weisheng and Lan, Kai and Zhou, Jianyi and Zhao, Bo and Su, Xiu and Tong, Junwen and Guan, Weili and Yang, Shuo},
  journal={arXiv preprint arXiv:2602.00557},
  year={2026}
}

@article{fu2026egograsp,
  title={Egograsp: World-space hand-object interaction estimation from egocentric videos},
  author={Fu, Hongming and Wang, Wenjia and Qiao, Xiaozhen and Potamias, Rolandos Alexandros and Komura, Taku and Yang, Shuo and Liu, Zheng and Zhao, Bo},
  journal={arXiv preprint arXiv:2601.01050},
  year={2026}
}

@article{brohan2023rt,
  title={Rt-2: Vision-language-action models transfer web knowledge to robotic control},
  author={Brohan, Anthony and Brown, Noah and Carbajal, Justice and Chebotar, Yevgen and Chen, Xi and Choromanski, Krzysztof and Ding, Tianli and Driess, Danny and Dubey, Avinava and Finn, Chelsea and others},
  journal={arXiv preprint arXiv:2307.15818},
  year={2023}
}

@inproceedings{kareer2025egomimic,
  title={Egomimic: Scaling imitation learning via egocentric video},
  author={Kareer, Simar and Patel, Dhruv and Punamiya, Ryan and Mathur, Pranay and Cheng, Shuo and Wang, Chen and Hoffman, Judy and Xu, Danfei},
  booktitle={2025 IEEE International Conference on Robotics and Automation (ICRA)},
  pages={13226--13233},
  year={2025},
  organization={IEEE}
}

@inproceedings{xie2026human2robot,
  title={Human2robot: Learning robot actions from paired human-robot videos},
  author={Xie, Sicheng and Cao, Haidong and Weng, Zejia and Xing, Zhen and Chen, Haoran and Shen, Shiwei and Leng, Jiaqi and Wu, Zuxuan and Jiang, Yu-Gang},
  booktitle={Proceedings of the AAAI Conference on Artificial Intelligence},
  volume={40},
  number={13},
  pages={11078--11086},
  year={2026}
}

@article{lepert2025phantom,
  title={Phantom: Training robots without robots using only human videos},
  author={Lepert, Marion and Fang, Jiaying and Bohg, Jeannette},
  journal={arXiv preprint arXiv:2503.00779},
  year={2025}
}

@article{lepert2025masquerade,
  title={Masquerade: Learning from in-the-wild human videos using data-editing},
  author={Lepert, Marion and Fang, Jiaying and Bohg, Jeannette},
  journal={arXiv preprint arXiv:2508.09976},
  year={2025}
}

@article{chi2025diffusion,
  title={Diffusion policy: Visuomotor policy learning via action diffusion},
  author={Chi, Cheng and Xu, Zhenjia and Feng, Siyuan and Cousineau, Eric and Du, Yilun and Burchfiel, Benjamin and Tedrake, Russ and Song, Shuran},
  journal={The International Journal of Robotics Research},
  volume={44},
  number={10-11},
  pages={1684--1704},
  year={2025},
  publisher={Sage Publications Sage UK: London, England}
}

@inproceedings{rombach2022high,
  title={High-resolution image synthesis with latent diffusion models},
  author={Rombach, Robin and Blattmann, Andreas and Lorenz, Dominik and Esser, Patrick and Ommer, Bj{\"o}rn},
  booktitle={2022 IEEE/CVF conference on computer vision and pattern recognition (CVPR)},
  pages={10674--10685},
  year={2022},
  organization={ieee}
}

@inproceedings{peebles2023scalable,
  title={Scalable diffusion models with transformers},
  author={Peebles, William and Xie, Saining},
  booktitle={2023 IEEE/CVF International Conference on Computer Vision (ICCV)},
  pages={4172--4182},
  year={2023},
  organization={IEEE}
}

@article{jiang2026cross,
  title={Cross-hand latent representation for vision-language-action models},
  author={Jiang, Guangqi and Liang, Yutong and Ye, Jianglong and Huang, Jia-Yang and Jing, Changwei and Duan, Rocky and Abbeel, Pieter and Wang, Xiaolong and Zou, Xueyan},
  journal={arXiv preprint arXiv:2603.10158},
  year={2026}
}

@inproceedings{wang2023mimicplay,
  title={MimicPlay: Long-Horizon Imitation Learning
         by Watching Human Play},
  author={Wang, Chen and Fan, Linxi and Sun, Jiankai
          and Zhang, Ruohan and Fei-Fei, Li and Xu, Danfei
          and Zhu, Yuke and Anandkumar, Anima},
  booktitle={Proceedings of the 7th Conference on Robot Learning},
  pages={201--221},
  volume={229},
  series={Proceedings of Machine Learning Research},
  publisher={PMLR},
  year={2023}
}

@inproceedings{qin2022dexmv,
  title={DexMV: Imitation Learning for Dexterous Manipulation
         from Human Videos},
  author={Qin, Yuzhe and Wu, Yueh-Hua and Liu, Shaowei
          and Jiang, Hanwen and Yang, Ruihan and Fu, Yang
          and Wang, Xiaolong},
  booktitle={European Conference on Computer Vision},
  pages={570--587},
  year={2022},
  publisher={Springer},
  doi={10.1007/978-3-031-19842-7_33}
}

@inproceedings{tao2025dexwild,
  title={DexWild: Dexterous Human Interactions
         for In-the-Wild Robot Policies},
  author={Tao, Tony and Srirama, Mohan Kumar
          and Liu, Jason Jingzhou and Shaw, Kenneth and Pathak, Deepak},
  booktitle={Proceedings of Robotics: Science and Systems},
  year={2025},
  doi={10.15607/RSS.2025.XXI.075}
}

@inproceedings{bu2025univla,
  title={Learning to Act Anywhere with Task-Centric Latent Actions},
  author={Bu, Qingwen and Yang, Yanting and Cai, Jisong
          and Gao, Shenyuan and Ren, Guanghui and Yao, Maoqing
          and Luo, Ping and Li, Hongyang},
  booktitle={Proceedings of Robotics: Science and Systems},
  year={2025},
  doi={10.15607/RSS.2025.XXI.014}
}

@inproceedings{chen2025moto,
  title={Moto: Latent Motion Token as the Bridging Language
         for Learning Robot Manipulation from Videos},
  author={Chen, Yi and Ge, Yuying and Tang, Weiliang
          and Li, Yizhuo and Ge, Yixiao and Ding, Mingyu
          and Shan, Ying and Liu, Xihui},
  booktitle={Proceedings of the IEEE/CVF International
             Conference on Computer Vision},
  pages={19752--19763},
  year={2025}
}

@inproceedings{ze2024dp3,
  title={3D Diffusion Policy: Generalizable Visuomotor Policy
         Learning via Simple 3D Representations},
  author={Ze, Yanjie and Zhang, Gu and Zhang, Kangning
          and Hu, Chenyuan and Wang, Muhan and Xu, Huazhe},
  booktitle={Proceedings of Robotics: Science and Systems},
  year={2024},
  doi={10.15607/RSS.2024.XX.067}
}

@inproceedings{zhang2025flowpolicy,
  title={FlowPolicy: Enabling Fast and Robust 3D Flow-Based Policy
         via Consistency Flow Matching for Robot Manipulation},
  author={Zhang, Qinglun and Liu, Zhen and Fan, Haoqiang
          and Liu, Guanghui and Zeng, Bing and Liu, Shuaicheng},
  booktitle={Proceedings of the AAAI Conference on Artificial Intelligence},
  volume={39},
  number={14},
  pages={14754--14762},
  year={2025},
  doi={10.1609/aaai.v39i14.33617}
}

@article{kingma2013auto,
    title={Auto-encoding variational bayes},
    author={Kingma, Diederik P and Welling, Max},
    journal={arXiv preprint arXiv:1312.6114},
    year={2013}
}

\clearpage
\appendix

\section{Implementation details}
\label{app:tokenizer_implementation}
\subsection{Action and geometry preprocessing}
\label{app:action_geometry}
\paragraph{Physical actions and current geometry.}
An OpenArm action concatenates a three-dimensional wrist position,
a six-dimensional rotation representation, and six Inspire-hand coordinates,
giving 15 values per timestep. Current hand geometry concatenates five
wrist-local fingertip positions with the wrist pose, giving 24 values. These
are different supervision targets: action learning uses future commanded wrist
poses and hand coordinates, whereas hand geometry-anchored alignment uses the currently
observed hand configuration. The bimanual wrist-hand representation
concatenates the two hands to give 30-dimensional actions and 48-dimensional
geometry. 
The single-hand architecture and action/geometry representations
are summarized in~\Cref{tab:implementation}.

\paragraph{Human demonstrations.}
Human videos are processed with a HaWoR-based reconstruction
pipeline~\citep{zhang2025hawor}. Reconstructed wrist motion is converted to the
robot coordinate convention, and hand articulation is retargeted to the shared
wrist-hand action representation~\citep{kim2026ego}. The reconstructed observed
hand also supplies current geometry targets. Robot demonstrations provide wrist
targets and hand commands directly, while robot geometry targets are obtained
from the observed configuration.

\begin{table}[h]
    \centering
    \caption{Architecture and representations.}
    \label{tab:implementation}

    \small
    \begin{tabular}{l l}
        \toprule
        Component & Configuration\\
        \midrule
        Backbone & RLDX-1-IMG~\citep{kim2026rldx}\\
        Policy inputs &
        \makecell[l]{One current egocentric RGB frame, proprioceptive state, \\ and instruction}\\
        Conditioning representation & 64 cognition tokens\\
        Physical action / current geometry & 15 / 24 coordinates per hand\\
        Data rate / action chunk & 20\,Hz / $K=16$ timesteps\\
        Latent action target & $K\times D_z$ posterior-mean coordinates ($K=16$, $D_z=64$)\\
        Geometry adapter & \makecell[l]{Token-wise residual MLP;\\hidden width 256; SiLU}\\
        Adapter output initialization & Zero projection weights and bias\\
        Geometry readout & Mean pooling followed by a linear map\\
        \makecell[l]{Tokenizer encoder /\\each decoder} & \makecell[l]{4 temporal blocks; width 256;\\4 attention heads}\\
        Default transfer decoder & Frozen human reconstruction decoder\\
        \bottomrule
    \end{tabular}
\end{table}

\paragraph{Normalization.}
Human retargeted actions are first converted to the canonical robot frame.
Tokenizer inputs are normalized coordinate-wise as
$\tilde a_j=\operatorname{clip}(2(a_j-a_j^{\min})/(a_j^{\max}-a_j^{\min})-1,-1,1)$,
where $a_j^{\min}$ and $a_j^{\max}$ are the minimum and maximum across the
human and robot source statistics.
The same rule applies separately to wrist position, the six rotation
components, and hand joint angles.
Both reconstruction decoders share these bounds and use
$\hat a_j=a_j^{\min}+\tfrac12(\operatorname{clip}(\hat{\tilde a}_j,-1,1)+1)
(a_j^{\max}-a_j^{\min})$.
Their outputs are already in the canonical Ego-pi-shared action frame, so the
human-to-robot conversion is not applied again.

Policy states are normalized independently using the same affine
formula, with lower and upper bounds given by the minimum source
1st percentile and maximum source 99th percentile, respectively.
These are shared bounds within the common embodiment tag, not
percentiles recomputed from pooled samples.

\subsection{Decoupled-decoder latent tokenizer architecture and objectives}
\label{app:tokenizer_objectives}
\paragraph{Shared encoder.}
The latent tokenizer projects each 15-dimensional action to width
256, adds fixed sinusoidal temporal embeddings, and applies four bidirectional
temporal Transformer blocks. Each block uses four attention heads of width 64,
pre-layer normalization, query/key normalization, and a SiLU-gated feed-forward
network. A final layer normalization and a $256\!\to\!64$ projection produce one
posterior-mean token per timestep. A $64\!\to\!64$ head predicts log-variance
from the mean, clipped to $[-8,8]$. No image, visual-feature target, or domain
token enters the encoder.

\paragraph{Embodiment-specific reconstruction decoders.}
Each complete decoder uses a $64\!\to\!256$ up-projection, sinusoidal positions, four Transformer
blocks of the same type as the encoder, and a non-affine final layer
normalization followed by a $256\!\to\!15$ output head. The human decoder is
initialized from the robot decoder before embodiment-specific reconstruction
training. The branches have independent temporal blocks and output heads,
not merely separate final linear layers.
Both tokenizers are trained from scratch for 100k steps:
Arm~I uses robot bottle/cup/doll/snack and human box/ball/towel
pick-and-place demonstrations;
Arm~II uses robot box/ball/towel pick-and-place and human
air-fryer opening/closing demonstrations.
Decoder copying occurs at random initialization, with no pretrained
tokenizer checkpoint or additional warm-start stage.
Both use SiLU-gated feed-forward layers with nominal expansion ratio; all dropout and stochastic depth are zero.

\paragraph{Reconstruction and KL reductions.}
For valid action coordinates, the tokenizer minimizes absolute reconstruction
error through the decoder selected by the example's source. Let
$\boldsymbol\mu_\psi(\mathbf A)$ and $\boldsymbol\ell_\psi(\mathbf A)$ denote
the posterior mean and log-variance produced by the shared encoder. The Gaussian
KL penalty sums $\tfrac12(\mu_j^2+\exp(\ell_j)-1-\ell_j)$ over latent
coordinates $j$. Reconstruction and cross-domain latent regularization use the
same reparameterized sample during tokenizer training.
The objective is
\begin{equation}
 \mathcal L_{\mathrm{tok}}=\mathcal L_{\mathrm{rec}}
 +10^{-6}\mathcal L_{\mathrm{KL}}
 +0.03\big(\mathcal L_{\mathrm{cent}}+\mathcal L_{\mathrm{var}}
                   +0.004\mathcal L_{\mathrm{cov}}\big).
 \label{eq:tokenizer_objective}
\end{equation}
Reconstruction is averaged over examples, all 16 temporal positions, and all 15 action coordinates; no coordinate padding is used, and end-of-episode repetitions remain included.
KL is averaged over examples and temporal positions but summed over the 64 latent dimensions. Human and robot examples receive equal per-example weight, without additional source balancing; their frequencies follow dataset-size-proportional sampling.

\paragraph{Position-wise centroid regression.}
For a minibatch containing complete action chunks from both sources, let
$\bar{\mathbf z}_{d,k}\in\mathbb R^{D_z}$ denote the mean sampled latent from
source $d\in\{\mathrm H,\mathrm R\}$ at relative chunk position $k$.
With $K=16$ and $D_z=64$, centroid regression is
\begin{equation}
 \mathcal L_{\mathrm{cent}}=\frac1{KD_z}
 \sum_{k=1}^{K}\|\bar{\mathbf z}_{\mathrm H,k}-\bar{\mathbf z}_{\mathrm R,k}\|_2^2.
 \label{eq:centroid}
\end{equation}
Relative chunk positions are not matched semantic stages. A valid bin requires
at least two human and two robot examples. Valid bins are weighted by the
smaller source count. With complete length-16 chunks, the reduction
is ~\Cref{eq:centroid}.

\paragraph{Variance and covariance regularization.}
Let $\mathbf X_d\in\mathbb R^{n_d\times D_z}$ stack latent tokens from
source $d$ after cross-device gathering. The variance term uses a population
variance reduction:
\begin{equation}
 \mathcal L_{\mathrm{var}}=\frac1{2D_z}
 \sum_{d\in\{\mathrm H,\mathrm R\}}\sum_{j=1}^{D_z}
 \max\big(0,1.75-\sqrt{\operatorname{Var}((\mathbf X_d)_{:,j})+10^{-4}}\big)^2.
 \label{eq:tokenizer_variance}
\end{equation}
Let
$\bar{\mathbf x}_d=n_d^{-1}\sum_{i=1}^{n_d}(\mathbf X_d)_{i,:}^{\top}$
and
$\widetilde{\mathbf X}_d=\mathbf X_d-\mathbf 1_{n_d}\bar{\mathbf x}_d^{\top}$.
The centered sample covariance is
$\boldsymbol\Sigma_d=\widetilde{\mathbf X}_d^{\top}\widetilde{\mathbf X}_d/(n_d-1)$,
and
\begin{equation}
 \mathcal L_{\mathrm{cov}}=\frac1{2D_z}
 \sum_{d\in\{\mathrm H,\mathrm R\}}\sum_{j\ne j'}(\boldsymbol\Sigma_d)_{j,j'}^2.
 \label{eq:tokenizer_covariance}
\end{equation}
Variance and covariance are computed globally within each source rather than
independently per semantic phase. These safeguards oppose collapse but do not
guarantee correct sample-level human-robot correspondence.

\paragraph{Frozen policy targets.}
After tokenizer training, the encoder and both decoders are frozen. Policy
targets are cached posterior means, not fresh VAE samples, and use no additional
latent standardization.

\subsection{Policy optimization and deployment}
\label{app:policy_optimization}
\label{app:deployment}
\paragraph{Hand geometry-anchored alignment.}
For conditioning tokens $\mathbf h_t$, the adapted representation is
$\mathbf h_t^{\mathrm{geo}}
=
\mathbf h_t+A_\phi(\mathbf h_t)$.
The linear readout $R_\omega$ acts on the mean-pooled representation,
while the action expert receives the full adapted token sequence. For sample $i$, let
$\delta_{i,j}\in\{0,1\}$ denote coordinate validity and let $\rho$ be the
unit-transition SmoothL1 penalty. The per-example loss is
\begin{equation}
 \ell_{\mathrm{geo}}(i)=
 \frac{\sum_j\delta_{i,j}\rho(\widehat g_{i,j}-g_{i,j})}
      {\max(1,\sum_j\delta_{i,j})}.
 \label{eq:geometry_loss}
\end{equation}
The batch loss $\mathcal L_{\mathrm{geo}}$ averages
$\ell_{\mathrm{geo}}(i)$ over examples with at least one valid coordinate.
The geometric readout is used only during training, with weight
$\lambda_{\mathrm{geo}}=0.2$ in the joint VLA objective.

\paragraph{Latent flow matching.}
We implement $\mathcal L_{\mathrm{pred}}$ using conditional flow
matching~\citep{lipman2022flow}. For a cached posterior-mean target
$\mathbf Z_t^d\in\mathbb R^{K\times D_z}$, let
$\boldsymbol\epsilon\in\mathbb R^{K\times D_z}$ have i.i.d. standard
normal entries, and let $\tau$ denote flow time sampled with the backbone's
training-time distribution. We define
\begin{equation}
 \mathbf Z_{t,\tau}^d
 =(1-\tau)\boldsymbol\epsilon+\tau\mathbf Z_t^d,
 \qquad
 \mathbf V_t^d
 =\mathbf Z_t^d-\boldsymbol\epsilon.
 \label{eq:latent_flow_path}
\end{equation}
The prediction objective is
\begin{equation}
 \mathcal L_{\mathrm{pred}}=
 \mathbb E\!\left[\frac{1}{KD_z}
 \|\mathbf v_\theta(\mathbf Z_{t,\tau}^d,
 \tau
 \mid
 \mathbf s_t,
 \mathbf h_t^{\mathrm{geo}})
       -\mathbf V_t^d\|_F^2\right],\qquad D_z=64.
 \label{eq:flow}
\end{equation}
Here $\mathbf v_\theta$ is the velocity predictor introduced in
\Cref{sec:preliminaries}, and $\|\cdot\|_F$ denotes the Frobenius norm
over the temporal and latent dimensions. The latent policy $\pi_\theta$ in
Eq.~\Cref{eq:deploy} is implemented by numerical integration of this field from
Gaussian noise. The shared
wrist-hand baseline and its geometry-grounded variant apply the prediction
objective to valid physical action coordinates instead of tokenizer latents.
We sample $u\sim\mathrm{Beta}(1.5,1)$ and set
$\tau=0.999(1-u)$, with no further time rescaling.
The backbone (RLDX-1-IMG) action expert uses four double-stream and eight single-stream blocks
with 24 attention heads of dimension 64.
Without the additional instruction pathway, the action expert receives
64 layer-normalized VLM latent tokens conditioned on the image and instruction;
the hand geometry-anchored baseline and \method{} additionally apply a residual
adapter with a 256-dimensional SiLU bottleneck.

Optimization settings and trainable modules are summarized
in~\Cref{tab:appendix_optimization,tab:appendix_trainable_modules}.

\begin{table}[h]
    \centering
    \caption{Optimization settings for tokenizer training, task-specific
    OpenArm fine-tuning (FT), and heterogeneous training (HT).}
    \label{tab:appendix_optimization}

    \small
    \begin{tabular}{l c c c}
        \toprule
        Setting & Tokenizer & OpenArm FT & HT\\
        \midrule
        Updates & 100k & 30k & 200k\\
        Global batch size & 512 & 64 & 128\\
        Optimizer & AdamW & AdamW & AdamW\\
        Learning rate & $5\times10^{-5}$ & $10^{-4}$ & $10^{-4}$\\
        Geometry-loss weight & Not applicable & 0.2 & 0.2\\
        \bottomrule
    \end{tabular}
\end{table}

\begin{table}[h]
    \centering
    \caption{Parameter updates during policy learning. Tokenizer training is a
    separate stage; its encoder and decoders are frozen during OpenArm FT.}
    \label{tab:appendix_trainable_modules}

    \small
    \begin{tabular}{l c c}
        \toprule
        Module & OpenArm FT & HT\\
        \midrule
        Vision encoder & Frozen & Frozen\\
        Language model & Frozen & Top four layers trainable\\
        Cognition-token parameters & Frozen & Trainable\\
        Geometry adapter / readout & Trainable / Trainable & Trainable / Trainable\\
        Tokenizer encoder / decoders & Frozen & Frozen where used\\
        \bottomrule
    \end{tabular}
\end{table}

\paragraph{Fine-tuning and data sampling.}
All OpenArm policies start from the same RLDX-1-IMG pretrained checkpoint
and are fine-tuned for 30k updates with global batch size 64.
Human-robot co-training uses equal domain weights and uniform task
weights within each domain; single-source baselines use uniform task
weights. Sampling cycles deterministically through action windows,
without enforcing an exact domain split in each minibatch.

Within-skill and object transfer use separately trained tokenizers,
each trained for 100k updates outside the policy-FT budget.
Tokenizer sampling is proportional to training-window counts.
The tokenizer recipe uses a fixed episode-level validation split;
policy FT uses no separate validation set.

\paragraph{Robot deployment.}
The decoder is fixed across target objects rather than switched per object.
We generate 16-action chunks using four explicit Euler steps.
Decoded actions are clipped to the tokenizer's normalized range
before inverse normalization. Wrist poses are converted to arm
commands using Mink IK with the DAQP solver and model-defined
joint limits, initialized from the observed joint state;
hand joint commands are applied directly.

\subsection{Additional instruction conditioning}
\label{app:instruction_implementation}
We evaluate instruction renewal as \method{} + Inst., separately
from the base \method{}. Instruction features are extracted from
the frozen VLM at text positions preceding the image, so they
encode the instruction without attending to visual tokens.
After trainable layer normalization, these features are
concatenated with the adapted vision-language cognition tokens
at the action-expert input.

Before every action-expert block, the instruction slots are
restored to their initial features, using the existing
stream projection where required, while the other token
representations evolve normally. We compare the full
instruction sequence with its last valid token; the input-only
control uses the full sequence without restoration. Padding
positions are masked throughout.

\subsection{Normalized human-robot latent gap}
\label{app:normalized_gap}

We measure the domain gap in the tokenizer latent space using the frozen
posterior-mean representation introduced in \Cref{sec:shared-action}.
After subsampling the two source domains to the same number $N$ of action
windows, let $\mathbf A_i^d$ denote the $i$-th selected action window from
source $d\in\{H,R\}$.
We encode each window as
\[
\mathbf Z_i^d
=
E_\psi(\mathbf A_i^d)
\in\mathbb R^{K\times D_z},
\]
where $E_\psi$ is the frozen posterior-mean mapping defined in
\Cref{sec:shared-action}.
For the $i$-th window, let
$\mathbf z_{i,k}^d\in\mathbb R^{D_z}$
denote the $k$-th temporal token of $\mathbf Z_i^d$,
with $k\in\{1,\ldots,K\}$.
Thus, every temporal token from every selected action window is treated as
one latent sample.

We define the domain-wise latent mean as
\begin{equation}
    \boldsymbol{\mu}_d
    =
    \frac{1}{NK}
    \sum_{i=1}^{N}
    \sum_{k=1}^{K}
    \mathbf z_{i,k}^d,
    \qquad d\in\{H,R\},
    \label{eq:latent_gap_mean}
\end{equation}
and the population variance of latent dimension $j$ as
\begin{equation}
    \sigma_{d,j}^{2}
    =
    \frac{1}{NK}
    \sum_{i=1}^{N}
    \sum_{k=1}^{K}
    \left(
        z_{i,k,j}^d-\mu_{d,j}
    \right)^2,
    \qquad
    j\in\{1,\ldots,D_z\},
    \label{eq:latent_gap_variance}
\end{equation}
where $z_{i,k,j}^d$ and $\mu_{d,j}$ denote the $j$-th coordinates of
$\mathbf z_{i,k}^d$ and $\boldsymbol{\mu}_d$, respectively.

We then compute the pooled per-dimension latent spread
\begin{equation}
    \bar{s}
    =
    \sqrt{
        \frac{1}{D_z}
        \sum_{j=1}^{D_z}
        \frac{
            \sigma_{H,j}^{2}
            +
            \sigma_{R,j}^{2}
        }{2}
    },
    \label{eq:latent_gap_scale}
\end{equation}
and define the normalized human--robot latent gap as
\begin{equation}
    G_{\mathrm{tok}}
    =
    \frac{
        \left\|
            \boldsymbol{\mu}_H
            -
            \boldsymbol{\mu}_R
        \right\|_2
    }{\bar{s}}.
    \label{eq:latent_gap}
\end{equation}
Here $D_z=64$ in all tokenizer experiments.
$G_{\mathrm{tok}}$ is a unitless measure of the Euclidean distance
between the human and robot latent centers relative to the pooled
per-dimension latent spread.

\section{Real-world data and evaluation}
\label{app:realworld_setup}
For controlled human-to-robot transfer experiments, we use an OpenArm bimanual platform equipped with two Inspire RH56F1 dexterous hands. RB-Y1 evaluation examines
robot-only downstream adaptation after heterogeneous training
(\Cref{app:rby1_adaptation}). The controlled transfer
experiments use only the right arm and hand. A head-mounted ZED 2i
camera provides $1280\times720$ RGB observations from a viewpoint
approximately matching the human demonstrations. Policies receive
one current right-camera RGB frame, a language instruction, and a
15-dimensional proprioceptive state comprising wrist position (3D),
wrist rotation (6D), and hand joint angles (6D). Wrist poses are
computed by forward kinematics in a fixed canonical camera frame. Rollouts run at 20Hz.

\subsection{Training data and baseline definitions}
\label{app:baseline_definitions}
\label{app:curriculum_data}
\paragraph{Transfer settings.}
Within-skill object transfer introduces Box, Ball, and Towel pick-and-place
through human demonstrations while retaining source robot pick-and-place
experience. Skill-level transfer introduces air-fryer opening and closing,
which are absent from the source robot demonstrations. The corresponding
robot in-domain evaluation tasks are Doll PnP and Box PnP, respectively.
Human and robot data are collected in the same environment, without frame-level human-robot correspondence.

\begin{table}[h]
    \centering
    \caption{Task and training-data conditions. PnP denotes pick-and-place.
    Policy FT uses all retained demonstrations without a held-out validation split;
    evaluation uses robot rollouts.}
    \label{tab:data_conditions}

    \small
    \begin{tabular}{l c c}
        \toprule
        Item & Object transfer & Skill transfer\\
        \midrule
        Human target tasks & Box / Ball / Towel PnP & Air-fryer Open / Close\\
        Robot in-domain evaluation & Doll PnP & Box PnP\\
        Exact robot training task list
        & Bottle / Cup / Doll / Snack PnP
        & Box / Ball / Towel PnP\\
        Retained robot episodes
        & 208 (64 / 48 / 32 / 64)
        & 300 (100 / 100 / 100)\\
        Retained human episodes per task
        & Box: 100; Ball: 100; Towel: 100
        & Open: 50; Close: 50\\
        \bottomrule
    \end{tabular}
\end{table}

\paragraph{Available corpus accounting.}
The recorded OpenArm corpus snapshot contains 908 episodes and approximately
208,600 frames at 20\,Hz. It comprises 508 robot pick-and-place episodes, 300 human
object-transfer episodes, and 100 human opening/closing episodes. The robot Open/Close
reference is a separate collection of 50 demonstrations per task.

\paragraph{Baseline scope.}
``Human-only'' and ``in-domain-only'' describe adaptation data, not the absence of prior knowledge in the common
robot-pretrained backbone. The target-robot-supervised Open/Close reference
used for the speed comparison is distinct from the source-task-only robot
baseline in ~\Cref{tab:main_object_transfer} and~\Cref{tab:main_skill_transfer}.

\begin{table}[h]
    \centering
    \caption{Baseline and variant definitions for the main comparison.}
    \label{tab:appendix_baselines}

    \small
    \renewcommand{\arraystretch}{1.2}
    \setlength{\tabcolsep}{4pt}

    \begin{tabular}{
        m{0.34\linewidth}
        m{0.17\linewidth}
        m{0.41\linewidth}
    }
        \toprule
        Condition
        & Adaptation sources
        & Action / conditioning interface\\
        \midrule

        Base policy human-only
        & Human targets
        & Human demonstrations only, using HaWoR-extracted fingertip and wrist
        motions as the action representation.\\

        Base policy in-domain-only
        & Source robot
        & Robot demonstrations only, using the native LeRobot robot action
        representation.\\

        Base policy co-train
        & \makecell[l]{Source robot\\+ human targets}
        & Human and robot data retain their respective action interfaces and use
        separate embodiment IDs; the target robot ID is used at inference.\\

        \midrule

        HaWoR-shared action~\citep{zhang2025hawor}
        & \makecell[l]{Source robot\\+ human targets}
        & Shared fingertip and wrist actions. Human actions come from HaWoR,
        while robot actions are mapped through forward kinematics; both use the
        same action decoder.\\

        Ego-Pi-shared action~\citep{kim2026ego}
        & \makecell[l]{Source robot\\+ human targets}
        & Shared physical wrist--hand actions without geometry grounding or a
        latent tokenizer.\\

        \midrule

        \makecell[l]{Ego-Pi-shared action~\citep{kim2026ego}\\
        + hand geometry-anchored alignment}
        & \makecell[l]{Source robot\\+ human targets}
        & Shared wrist--hand actions with our hand geometry-anchored alignment.\\

        \method{}
        & \makecell[l]{Source robot\\+ human targets}
        & Hand geometry-anchored alignment with a decoupled-decoder latent
        tokenizer.\\

        \method{} + Inst.
        & \makecell[l]{Source robot\\+ human targets}
        & \method{} with the additional instruction-renewal pathway.\\

        \bottomrule
    \end{tabular}

\end{table}

\subsection{Task-progress scoring}
\label{app:evaluation_protocol}
We report a weighted task score to capture partial execution as well as full
completion. For $n$ rollouts and the highest progress level $w_i$ reached by
rollout $i$,
\begin{equation}
 S_{\mathrm w}=\frac{100}{n}\sum_{i=1}^{n}w_i.
 \label{eq:weighted_score}
\end{equation}
Pick-and-place uses levels 0 for failure, 0.3 for reach, 0.6 for grasp, 0.8 for
moving the object, and 1.0 for full completion. Opening and closing use levels on the same scale, with task-specific criteria.
Task-specific progress milestones are detailed
in~\Cref{tab:appendix_scoring_rubric}.

\begin{table}[h]
    \centering
    \caption{Task-progress weights and milestones.}
    \label{tab:appendix_scoring_rubric}

    \small
    \begin{tabular}{c l l l}
        \toprule
        $w_i$ & Pick-and-place & Open & Close\\
        \midrule
        0.0 & Failure & Failure & Failure\\
        0.3 & Reach to object & Reach to handle & Reach to handle\\
        0.6 & Valid grasping of object & Total grasp of handle & Total grasp of handle\\
        0.8 & The object lifted up  & Opened halfway & Closed halfway\\
        1.0 & Complete task & Complete task & Complete task \\
        \bottomrule
    \end{tabular}
\end{table}

\subsection{ID/OOD placement and rollout protocol}
\label{app:placement_protocol}
For each target object, the object-transfer evaluation contains 20
in-distribution (ID) and 10 out-of-distribution (OOD) placement trials, totaling
90 target-task rollouts per policy. Skill transfer contains 15 Open and 15
Close trials, totaling 30. Robot in-domain performance is evaluated separately
with 20 trials on the corresponding source task.
Per-task evaluation counts are summarized
in~\Cref{tab:appendix_rollout_counts}.

\begin{table}[h]
    \centering
    \caption{Evaluation counts per policy in the main transfer comparison. The
    robot in-domain trials are separate from the target-task averages.}
    \label{tab:appendix_rollout_counts}

    \small
    \renewcommand{\arraystretch}{1.10}
    \setlength{\tabcolsep}{6pt}

    \begin{tabular}{l c c c}
        \toprule
        Evaluation group & ID & OOD & Total\\
        \midrule

        \multicolumn{4}{l}{\textbf{I: Object transfer}}\\
        \addlinespace[2pt]

        Object target: Box PnP
        & 20 & 10 & 30\\

        Object target: Ball PnP
        & 20 & 10 & 30\\

        Object target: Towel PnP
        & 20 & 10 & 30\\

        \cmidrule(lr){1-4}
        \textbf{Object target subtotal}
        & \textbf{60} & \textbf{30} & \textbf{90}\\

        \addlinespace[2pt]
        \makecell[l]{Robot in-domain: Doll PnP}
        & \multicolumn{2}{c}{\makecell[c]{Separate source-task evaluation}}
        & 20\\

        \midrule

        \multicolumn{4}{l}{\textbf{II: Skill transfer}}\\
        \addlinespace[2pt]

        Skill target: Open
        & \multicolumn{2}{c}{No ID/OOD split}
        & 15\\

        Skill target: Close
        & \multicolumn{2}{c}{No ID/OOD split}
        & 15\\

        \cmidrule(lr){1-4}
        \textbf{Skill target subtotal}
        & \multicolumn{2}{c}{\textbf{Open + Close}}
        & \textbf{30}\\

        \addlinespace[2pt]
        \makecell[l]{Robot in-domain: Box PnP}
        & \multicolumn{2}{c}{\makecell[c]{Separate source-task evaluation}}
        & 20\\

        \bottomrule
    \end{tabular}
\end{table}

\paragraph{Placement distribution and evaluation protocol.}

\begin{figure}[t]
    \centering

    \newlength{\panelheight}
    \setlength{\panelheight}{0.33\columnwidth}

    % ------------------------
    % (a) Robot setup
    % ------------------------
    \begin{minipage}[t]{0.37\linewidth}
        \centering
        \vspace{0pt}
        \includegraphics[
            height=\panelheight
        ]{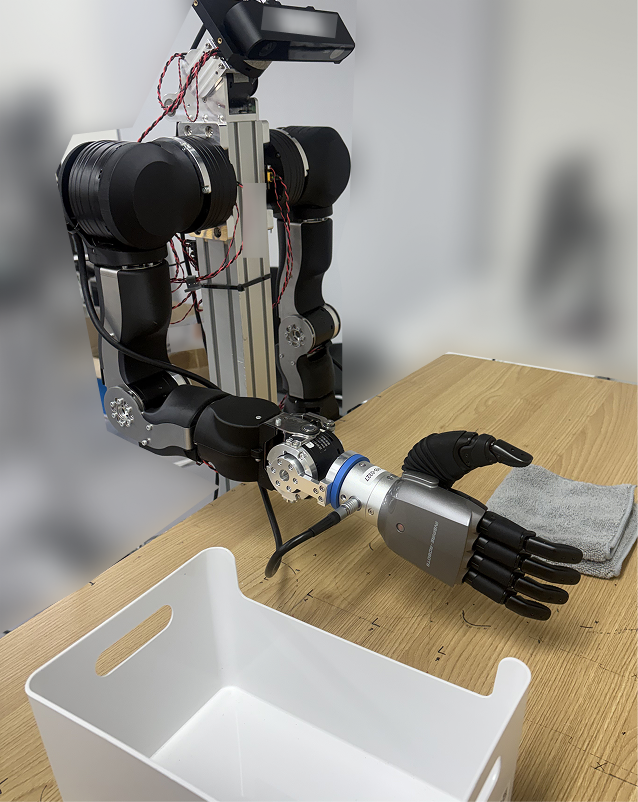}

        \vspace{2pt}
        \textbf{(a) OpenArm setup}
    \end{minipage}
    \hfill
    % ------------------------
    % (b) Evaluation distributions
    % ------------------------
    \begin{minipage}[t]{0.59\linewidth}
        \centering
        \vspace{0pt}
        \IfFileExists{figures/placement.png}{
            \includegraphics[
                height=\panelheight
            ]{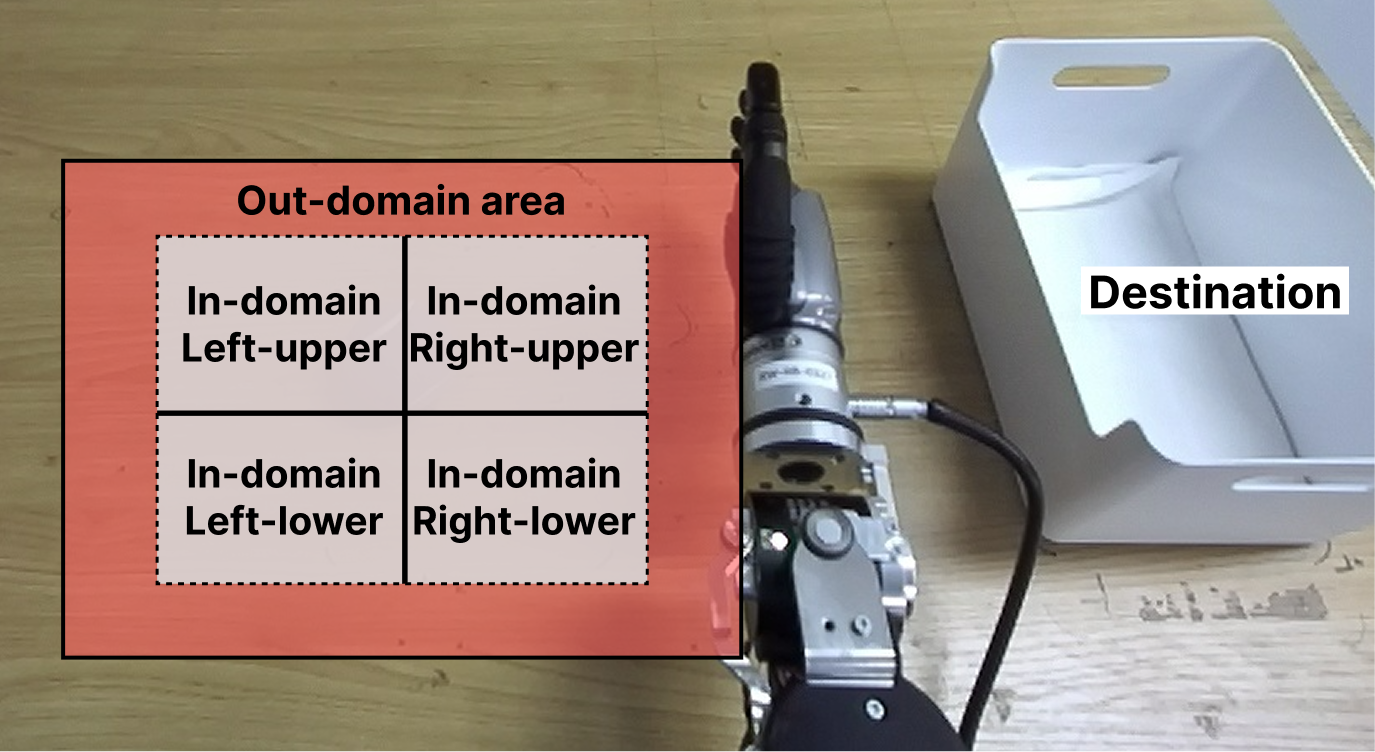}
        }

        \vspace{2pt}
        %\captionsetup{width=3\linewidth}
        \textbf{(b) Within-skill object transfer placement regions}
    \end{minipage}

    \caption{
        \textbf{OpenArm robot setup and evaluation distributions.}
        \textbf{(a)} OpenArm robot platform used for real-world evaluation.
        \textbf{(b)} For object transfer, in-distribution (ID) rollouts are sampled
        from four workspace quadrants (right-upper, right-lower, left-upper, and
        left-lower), with five rollouts per quadrant and randomized object orientation.
        Out-of-distribution (OOD) rollouts are sampled outside the ID placement region (red area),
        again with randomized orientation.
        For skill transfer, the initial openness of the air fryer is randomized for all 30 trials.
    }
    \label{fig:placement_distribution}
\end{figure}

For object transfer, we divide the nominal workspace into four quadrants-right-upper (RU), right-lower (RD), left-upper (LU), and left-lower (LD)-and sample five initial object placements from each quadrant, yielding 20 in-distribution (ID) rollouts per object. Object orientation is randomized independently for each rollout. We use the same quadrant-based placement protocol during human data collection, such that the ID evaluation distribution follows the spatial support observed during training. For out-of-distribution (OOD) evaluation, we instead sample 10 initial placements outside these quadrants, while continuing to randomize object orientation. For skill transfer, the initial state of the air fryer is randomized by varying its degree of opening for both the Open and Close tasks.
~\Cref{fig:placement_distribution} illustrates the corresponding ID and OOD placement regions.

\subsection{Evaluation aggregation and accounting}
\label{app:evaluation_accounting}
\label{app:evidence}

Because Box, Ball, and Towel use equal numbers of trials, the reported target-task average is the mean of the three per-object scores and is equivalently the weighted task score pooled over all 90 target-task rollouts. The corresponding pooled ID/OOD score is
\[
S_{\mathrm w}^{\mathrm{all}}
=
\frac{60S_{\mathrm w}^{\mathrm{ID}}+
30S_{\mathrm w}^{\mathrm{OOD}}}{90}.
\]
For skill transfer, Open and Close are each evaluated with 15 rollouts, so their reported average is equivalently the score pooled over all 30 target-task rollouts. The separate 20-rollout robot in-domain evaluation is not included in either target-task average.

All aggregate scores are computed from the per-rollout progress annotations before rounding; table entries are rounded to one decimal place.

Each reported condition uses one independently trained policy.
All scheduled evaluation rollouts are included in the reported scores;
failed task executions remain failures and are not retried or excluded.

\section{Additional results and ablations}
\label{app:additional_transfer}

\subsection{Cumulative task progress}
\label{app:cumulative_progress}
\Cref{tab:cumulative_progress} reports cumulative target-task progress
for the shared wrist-hand baseline and \method{}. Thresholds refer to the
rubric in ~\Cref{app:evaluation_protocol}. The weighted
scores are the corresponding main-table values.

\begin{table}[h]
    \centering
    \caption{Cumulative target-task progress (\%) and weighted task scores
    (0-100). $R_{\geq0.3}$, $R_{\geq0.6}$, $R_{\geq0.8}$, and $R_{=1.0}$ denote
    increasing completion thresholds; the last is full-task success.}
    \label{tab:cumulative_progress}

    \small
    \begin{tabular}{l c c c c c c}
        \toprule
        Policy & $n$ & $R_{\geq0.3}$ & $R_{\geq0.6}$ & $R_{\geq0.8}$ & $R_{=1.0}$ & $S_{\mathrm w}$\\
        \midrule
        \multicolumn{7}{l}{\textbf{I: Object transfer}}\\
        Ego-Pi-shared action & 90 & 68.9 & 12.2 & 12.2 & 2.2 & 27.2\\
        \method{} & 90 & 95.6 & 46.7 & 33.3 & 12.2 & 51.8\\
        \method{} + HT & 90 & 96.7 & 60.0 & 30.0 & 21.1 & 57.2\\
        \midrule
        \multicolumn{7}{l}{\textbf{II: Skill transfer}}\\
        Ego-Pi-shared action & 30 & 0.0 & 0.0 & 0.0 & 0.0 & 0.0\\
        \method{} & 30 & 100.0 & 66.7 & 33.3 & 30.0 & 62.7\\
        \method{} + HT & 30 & 100.0 & 83.3 & 76.7 & 56.7 & 81.7\\
        \bottomrule
    \end{tabular}
\end{table}

\subsection{Hand geometry-anchored alignment under placement shift}
\label{app:geometry_shift}
\Cref{tab:geometry_grounding_ablation} retains the intermediate
hand geometry-anchored alignment with physical-action condition, in addition to the shared-action baseline and full method. ID and OOD counts follow ~\Cref{app:placement_protocol}.

\begin{table}[h]
    \centering
    \caption{
    Effect of geometry grounding on object transfer.
    Robot in-domain evaluates the familiar Doll PnP task, while human transfer
    reports weighted task scores under in-distribution (ID) and
    out-of-distribution (OOD) object placements.
    All values are weighted task scores (0-100).
    }
    \label{tab:geometry_grounding_ablation}

    \small

    \begin{tabular}{l c c c c}
        \toprule
        & \multicolumn{1}{c}{Robot in-domain}
        & \multicolumn{3}{c}{Human transfer} \\
        \cmidrule(lr){2-2}
        \cmidrule(lr){3-5}

        Method
        & \makecell{Doll PnP}
        & \makecell{ID}
        & \makecell{OOD}
        & \makecell{Avg.}
        \\
        \midrule

        \makecell[l]{Base policy co-train \\
        + Ego-Pi-shared action~\citep{kim2026ego}}
        & 30.0
        & 30.2
        & 21.3
        & 27.2
        \\

        \makecell[l]{
            + Ego-Pi-shared action~\citep{kim2026ego} \\
            \phantom{+ }Hand geometry-anchored alignment
            }
        & 47.5
        & 36.0
        & 32.0
        & 34.7
        \\

        \textbf{+ \method{} (Ours)}
        & \textbf{49.0}
        & \textbf{55.0}
        & \textbf{45.3}
        & \textbf{51.8}
        \\

        \bottomrule
    \end{tabular}
\end{table}

\subsection{Decoder selection ablations}
\label{app:decoder_tokenizer_ablations}
\paragraph{Inference decoder selection.}
Both decoder branches are evaluated using the same within-skill object-transfer
protocol described in ~\Cref{app:placement_protocol}, with 30 rollouts
per object and the same ID/OOD evaluation split.
Only the reconstruction branch used at inference is changed; the latent policy
and all other evaluation settings remain fixed. As shown in ~\Cref{tab:decoder_ablation}, the human decoder trained on human action data achieves better transfer performance than the robot decoder.
The human decoder is therefore used for all main transfer results, without per-object
decoder switching.

\begin{table}[h]
    \centering
    \caption{Inference-decoder selection for the within-skill object transfer latent
    policy. Values are weighted task scores (0-100).}
    \label{tab:decoder_ablation}

    \small
    \begin{tabular}{l c c c c}
        \toprule
        Reconstruction decoder & Box & Ball & Towel & Avg.\\
        \midrule
        Robot decoder & 37.0 & 16.0 & 45.0 & 32.7\\
        Human decoder & \textbf{53.0} & \textbf{49.0} & \textbf{53.3} & \textbf{51.8}\\
        \bottomrule
    \end{tabular}
\end{table}

\begin{figure*}[t]
    \centering
    \includegraphics[width=\textwidth]{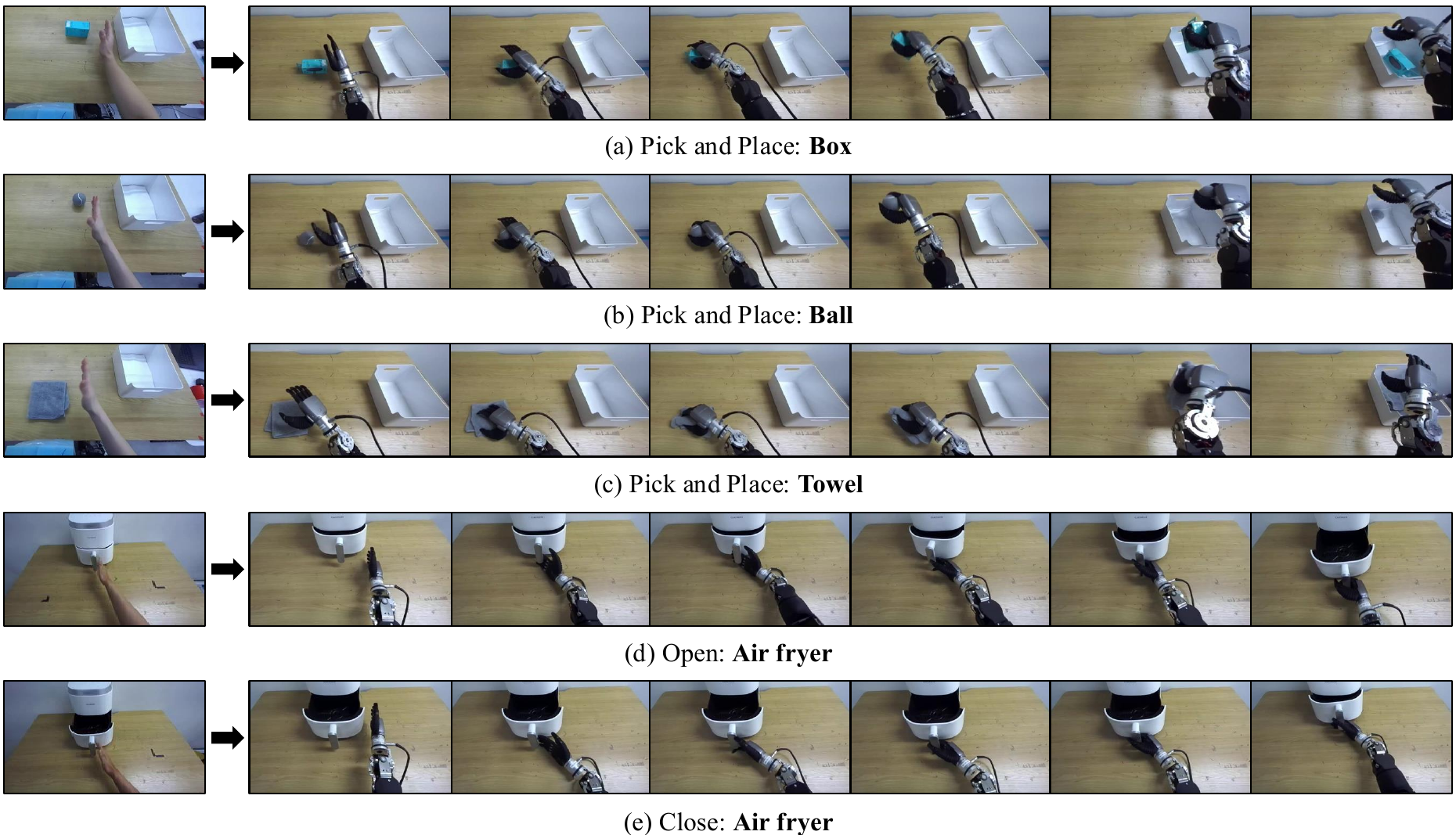}
    \caption{
    \textbf{Qualitative transferred-task rollouts.}
    Representative successful \method{} executions on box, ball, towel, open, and close, spanning both object-level and skill-level transfer.
    }
    \label{fig:qualitative-success}
\end{figure*}

\subsection{Execution dynamics}
To compare source-specific motion dynamics, we additionally train a
robot-supervised reference on 50 Open and 50 Close demonstrations.
As reported in ~\Cref{tab:action-tokenizer-decoding}, the
human-transfer policy reaches an average wrist speed of 7.36\,cm/s,
compared with 2.52\,cm/s for the robot-supervised reference.
The two policies are trained separately; this comparison is not an
inference-decoder swap. Wrist speed is computed from consecutive policy-predicted wrist positions at 20\,Hz over each full successful rollout, including stationary periods, then averaged across episodes within each task and equally across Open and Close.

\section{Heterogeneous multi-embodiment training (HT)}
\label{app:midtraining}
The VLA backbone and
hand geometry-anchored alignment are shared, while group-specific state and
action interfaces accommodate different embodiments and action representations.

\subsection{Data mixture and action interfaces}
\label{app:midtraining_data}
The mixture contains 14,548 episodes and 5,014,530 frames.
In-house ALLEX robot data contain 480 demonstrations. In-house human data
combine 1,480 head-mounted camera demonstrations with 930 demonstrations recorded by OpenArm ZED 2i camera. Single-hand OpenArm combines robot and human
pick-and-place data with human air-fryer opening/closing demonstrations;
EgoDex and bimanual OpenArm provide the remaining data. The ALLEX collections
use 30\,Hz, and all other collections use 20\,Hz. Overall source combination is described in ~\Cref{tab:appendix_mt_mixture}.

\begin{table}[h]
    \centering
    \caption{Heterogeneous training data by collection source.}
    \label{tab:appendix_mt_mixture}

    \small
    \begin{tabular}{l c c}
        \toprule
        Data source & Episodes & Frames\\
        \midrule
        EgoDex~\citep{hoque2026egodex} & 10,000 & 1,714,342\\
        In-house human data & 2,410 & 1,754,308 \\
        Single-hand OpenArm & 1,035 & 236,751\\
        Bimanual OpenArm & 623 & 249,488\\
        In-house ALLEX robot data & 480 & 1,059,641\\
        \midrule
        Total & 14,548 & 5,014,530\\
        \bottomrule
    \end{tabular}
\end{table}

We initialize five new state/action MLP interfaces while retaining the
36 pretrained interfaces of RLDX-1-IMG. In-house ALLEX robot and in-house human data share the
50D action interface and its frozen tokenizer; single-hand OpenArm uses a
separate 15D tokenizer. EgoDex and bimanual OpenArm predict native
30D wrist-hand actions, normalized using group-specific first and 99th
percentiles with invalid coordinates masked. Tokenizer-based interfaces
retain their own action normalization. All interfaces use 16-step action
chunks and current egocentric images; ALLEX uses stereo views, whereas the
other collections use a single view.

\subsection{Optimization and downstream adaptation}
\label{app:midtraining_optimization}
State dropout is 0.3, and training images receive color jitter with
brightness, contrast, and saturation strength 0.6 and hue strength 0.16.

\paragraph{Shared hand geometry-anchored alignment.}
A residual adapter of hidden width 256 and a shared linear readout predict
48-dimensional wrist-fingertip geometry, with 24 coordinates per hand.
ALLEX targets are obtained through forward kinematics; the single-hand
OpenArm group supervises only the right-hand coordinates and masks the left
hand. The remaining groups use observed bimanual geometry. Geometry grounding
has weight 0.2 and uses the loss in Eq.~\Cref{eq:geometry_loss},
regardless of whether the action target is latent or native. The backbone's
auxiliary physics-loss coefficient is 0.1.

For downstream evaluation, both the original and HT backbones are
fine-tuned on the same OpenArm object- and skill-transfer tasks using the
protocol in ~\Cref{app:realworld_setup}.

\subsection{Task-level transfer results}
\label{app:openarm_midtraining}
\Cref{tab:mt_object_transfer} and \Cref{tab:mt_skill_transfer} report weighted task scores. This
comparison measures the benefit of additional HT, rather than
isolating hand geometry-anchored alignment within a matched MT recipe.

% ------------------------
% I: Object transfer
% ------------------------
\begin{table}[h]
\centering
\small
\caption{
Effect of heterogeneous training on object transfer.
Robot in-domain reports performance on a robot-data-trained task, while
Human transfer evaluates target tasks introduced through human demonstrations.
Each entry is the weighted task score (0-100).
}
\label{tab:mt_object_transfer}

\renewcommand{\arraystretch}{1.08}

\begin{tabular}{l c c c c c}
    \toprule
    & \multicolumn{1}{c}{Robot in-domain}
    & \multicolumn{4}{c}{Human transfer} \\
    \cmidrule(lr){2-2}
    \cmidrule(lr){3-6}
    Method
    &
    \makecell{Doll}
    &
    \makecell{Box}
    &
    \makecell{Ball}
    &
    \makecell{Towel}
    &
    \makecell{Avg.}
    \\

    \midrule

    \method{}
    & 49.0
    & 53.0
    & 49.0
    & 53.3
    & 51.8 \\

    \rowcolor{gray!10}
    \textbf{HT + \method{}}
    & \textbf{69.5}
    & \textbf{56.7}
    & \textbf{53.3}
    & \textbf{61.7}
    & \textbf{57.2} \\

    \bottomrule
\end{tabular}
\end{table}

% ------------------------
% II: Skill transfer
% ------------------------
\begin{table}[h]
\centering
\small
\caption{
Effect of heterogeneous training on skill transfer.
Robot in-domain reports performance on a robot-data-trained task, while
Human transfer evaluates target tasks introduced through human demonstrations.
Each entry is the weighted task score (0-100).
}
\label{tab:mt_skill_transfer}

\renewcommand{\arraystretch}{1.08}

\begin{tabular}{l c c c c}
    \toprule
    & \multicolumn{1}{c}{Robot in-domain}
    & \multicolumn{3}{c}{Human transfer} \\
    \cmidrule(lr){2-2}
    \cmidrule(lr){3-5}
    Method
    &
    \makecell{Box PnP}
    &
    \makecell{Open}
    &
    \makecell{Close}
    &
    \makecell{Avg.}
    \\

    \midrule

    \method{}
    & \textbf{81.0}
    & 50.7
    & 74.7
    & 62.7 \\

    \rowcolor{gray!10}
    \textbf{HT + \method{}}
    & \underline{80.5}
    & \textbf{64.7}
    & \textbf{98.7}
    & \textbf{81.7} \\

    \bottomrule
\end{tabular}

\end{table}

\subsection{RB-Y1 robot-only adaptation}
\label{app:rby1_adaptation}

\begin{figure}[t]
    \centering

    \begin{subfigure}[t]{0.48\linewidth}
        \centering
        \includegraphics[
            height=0.7\columnwidth,
            keepaspectratio
        ]{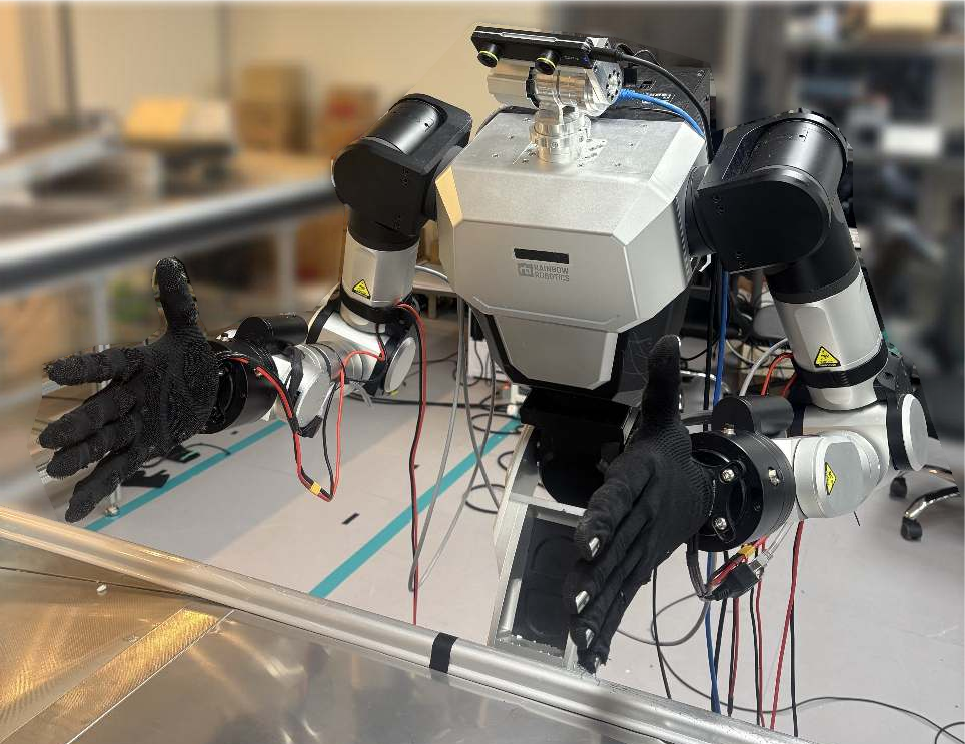}
        \caption{RB-Y1 robot setup}
        \label{fig:rby1_setup-a}
    \end{subfigure}
    \hfill
    \begin{subfigure}[t]{0.48\linewidth}
        \centering
        \includegraphics[
            height=0.7\columnwidth,
            keepaspectratio
        ]{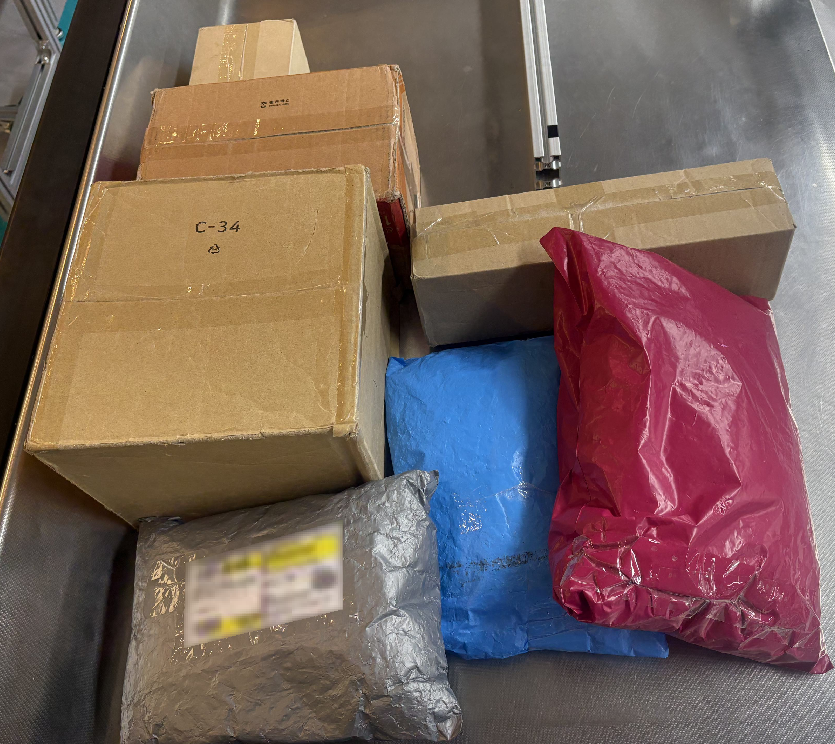}
        \caption{Evaluation objects}
        \label{fig:rby1_setup-b}
    \end{subfigure}

    \caption{
        \textbf{RB-Y1 experimental setup.}
        (a) Bimanual robot platform.
        (b) Boxes and polybags used for parcel handling.
    }
    \label{fig:rby1_setup}
\end{figure}

\paragraph{Hardware and observations.}
We evaluate robot-only adaptation on the RB-Y1
bimanual platform, equipped with two 7-DoF arms and two
20-DoF Wuji Hand 2 dexterous hands, as shown in~\Cref{fig:rby1_setup-a}.
A head-mounted ZED Mini stereo camera provides two RGB
observations (left and right), each resized to $256\times256$.
The policy receives both current RGB frames and a
58-dimensional proprioceptive state comprising the left and
right wrist poses (18D) and hand joint angles (40D).
Each wrist pose consists of a 3D position and a 6D rotation
representation. The policy predicts 58D absolute actions
in the robot-base coordinate convention, with absolute
joint-angle commands for both hands.

\paragraph{Robot-only fine-tuning.}
All downstream policies are fine-tuned on the same robot-only
dataset containing 726,752 frames after temporal retiming.
The dataset and action targets are sampled at 30\,Hz,
with an action chunk length of 16 and 30k fine-tuning updates.
We compare three HT variants initialized from the same pretrained backbone: robot-only HT, human-robot HT, and geometry-grounded human-robot HT. Robot-only HT uses the same training mixture with human demonstrations excluded, while the two human-robot variants differ only in geometry alignment. All variants use the same optimization settings for 200k HT updates and subsequently undergo robot-only downstream fine-tuning without the shared latent action tokenizer.

\paragraph{Evaluation protocol.}
The task requires orienting parcels (\Cref{fig:rby1_setup-b}), including boxes and
polybags, such that their shipping labels face upward,
and transferring them onto a conveyor belt.
We evaluate ten distinct objects with randomized initial
orientations, repeating each object five times for a total
of 50 trials per policy.
Success is binary: a trial receives a score of 1 only if
the parcel is placed on the conveyor belt with its shipping
label facing upward, and 0 otherwise.
No intermediate subtask scores or partial progress
credits are assigned.

\end{document}